\documentclass[conference]{IEEEtran}

\usepackage{cite}

\ifCLASSINFOpdf
\usepackage[pdftex]{graphicx}
\graphicspath{./}
\DeclareGraphicsExtensions{.pdf,.jpeg,.png}
\else
\usepackage[dvips]{graphicx}
\graphicspath{{figures/eps/}}
\DeclareGraphicsExtensions{.eps}
\fi
\usepackage{multirow}

\usepackage[cmex10]{amsmath}
\usepackage{textcomp}
\usepackage{color,soul}

\usepackage{array}
\usepackage{amssymb}

\usepackage{mdwmath}
\usepackage{mdwtab}

\usepackage{eqparbox}

\usepackage[tight,footnotesize]{subfigure}

\usepackage{stfloats}

\usepackage{url}

\usepackage{color}  
\newcommand{\awd}[1]{}               
\newcommand{\awn}[1]{{#1}}         
\usepackage{fancyhdr}
\fancypagestyle{firstP}{
	\fancyhf{}

	\fancyhead[L]{IEEE UPINLBS'2016\\Nov. 3\texttildelow4, 2016, Shanghai, P.R. China}
	\fancyfoot[L]{978-1-5090-2879-5/16/\$31.00 \textcopyright2016 IEEE}
	\fancyfoot[R]{\thepage}
}

\usepackage{lipsum}
\begin{document}
	%
	\title{Application of backpropagation neural networks to both stages of fingerprinting based WIPS}
	
	\author{\IEEEauthorblockN{Caifa Zhou, Andreas Wieser}
		\IEEEauthorblockA{ETH Zurich, IGP\\
			Stefano-Franscini-Platz 5, 8093 Zurich, Switzerland\\
			Email: \{caifa.zhou@geod.baug.ethz.ch; andreas.wieser@geod.baug.ethz.ch\}}
	}

	

	\maketitle
	\thispagestyle{firstP}
	
	\begin{abstract}
		We propose a scheme to employ backpropagation neural networks (BPNNs) for both stages of fingerprinting-based indoor positioning using WLAN/WiFi signal strengths (FWIPS): radio map construction during the offline stage, and localization during the online stage. Given a training radio map (TRM), i.e., a set of coordinate vectors and associated WLAN/WiFi signal strengths of the available access points, a BPNN can be trained to output the expected signal strengths for any input position within the region of interest (BPNN-RM). This can be used to provide a continuous representation of the radio map and to filter, densify or decimate a discrete radio map. Correspondingly, the TRM can also be used to train another BPNN to output the expected position within the region of interest for any input vector of recorded signal strengths and thus carry out localization (BPNN-LA).
		Key aspects of the design of such artificial neural networks for a specific application are the selection of design parameters like the number of hidden layers and nodes within the network, and the training procedure. Summarizing extensive numerical simulations, based on real measurements in a testbed, we analyze the impact of these design choices on the performance of the BPNN and compare the results in particular to those obtained using the $k$ nearest neighbors ($k$NN) and weighted $k$ nearest neighbors approaches to FWIPS.		
		The results indicate that BPNN-RM can help to reduce the workload for radio map generation significantly by allowing to sample the signal strengths at significantly less positions during the offline phase while still obtaining equal or even slightly better accuracy during the online stage as when directly applying the sampled radio map to (weighted) $k$NN. In the scenario analyzed within the paper the workload can be reduced by almost 90\%. We also show that a BPNN-LA with only 1 hidden layer outperforms networks with more hidden layers and yields positioning accuracy comparable to or even slightly better than $k$NN but with less computational burden during the online stage.
	\end{abstract}
	
	

	%
	
	\section{Introduction}\label{intro}
As \awn{key requirements} of context awareness and pervasive computing, indoor location based services (ILBSs) as well as the systems to provide indoor positioning \awn{have} attract\awn{ed} much attention from both \awn{academia} and industry over \awn{the last} two decades \cite{Abowd}. \awn{Their} predicted market value is up to 2.5\awn{bn} dollars by 2020 \cite{wifi_retail}. Various indoor positioning systems (IPSs) based on different signals, for instance WLAN/WiFi \cite{Padmanabhan2000}, Bluetooth, radio frequency identification (RFID) \cite{6934184}, light \cite{liu2007survey}, magnetic field \cite{6418880}, ultra-wide band (UWB) \cite{gigl2007analysis} and ultrasound/acoustic sound \cite{hazas2006broadband,mandal2005beep}, have been investigated as alternatives to global navigation satellites systems (GNSSs) which are unavailable or too inaccurate in the indoor environment \cite{He2016}.
	
The application of WLAN/WiFi signals has attracted continuous attention due to the widespread deployment of WLANs and \awn{availability of} WiFi enabled mobile devices. From this perspective, WLAN/WiFi based IPSs (WIPSs) are cost-effective because they often do not require any additional infrastructure and no specific hardware for the purpose of positioning. Fingerprinting based localization is a very promising positioning approach for IPSs because it also works if there is no line-of-sight (LoS) signal propagation between the access points (APs) and the receivers. Methods based on trilateration and triangulation depend on the availability of LoS signal\awn{s} and are negatively affected by \awn{non-}LoS signal propagation which is common within buildings. In this paper, the authors \awn{thus} focus on fingerprinting based WIPS (FWIPS). 
	
Generally, \awn{a} FWIPS involves two stages: a site survey offline stage and a user positioning online stage. In the offline stage, the site survey is conducted to create the radio map (RM) which represents the expected WLAN/WiFi signal strength for all locations within the region of interest (RoI). Often the survey consists of sampling the received signal strength (RSS) from all visible APs at given reference points (RPs) with known locations \awn{with}in the RoI. The collection of all RPs and the corresponding RSS vectors is stored in a fingerprinting database. The raw data in the fingerprinting database are then converted into the radio map which is used for online positioning. Here in this paper the original RM is just a set of RP coordinates \awn{and} the respective measured RSS values without any mapping or filtering. During the online stage, a user measures the RSS vector and matches it to the RM using some defined similarity metric in the signal space (e.g.\awn{,} Euclidean distance) under the general assumption that the location of the user is embedded in the readings of RSS. In a simple approach, the \awn{points} whose RSS within the RM are the most "similar" \awn{ones} to the user's RSS \awn{values} are utilized to estimate the user location (e.g.\awn{,} using the $k$ nearest neighbor ($k$NN) algorithm). The main bottleneck which constrains the widespread commercial application of FWIPS is the heavy workload to build the RM for a large area (e.g.\awn{,} an entire airport or a big mall) and to keep the RM up-to-date \cite{He2016}.
	
Apart from the \awn{separate and time-consuming} manual collection of \awn{RSS values at known positions} two \awn{other} methods \awn{for obtaining the RM are available:} unsupervised fingerprinting and partial fingerprinting \cite{7362027}. \awn{For the} unsupervised fingerprinting  the RM \awn{is created by} employing \awn{an} indoor propagation model of radio \awn{waves} to predict the RSS values \awn{with}in the RoI. \awn{This} requires accurate information about the structure (e.g.\awn{,} floors, walls, windows and doors) of \awn{the} building, the materials (e.g.\awn{,}  concrete, wood, metal and glass) used for the respective structures as well as the position \awn{and} configuration of \awn{the} APs (e.g.\awn{,} power, gain of the antenna and protocols). This approach \awn{thus requires a labor intensive site survey or detailed building plans and assumptions, and it yields} bad performance in case of \awn{invalid assumptions or changes of any of the parameters}.
	
\awn{P}artial fingerprinting utilizes crowdsourcing \awn{t}o improve the efficiency of RM construction and update \cite{He2016, Park2010, Wu2013}. Depending on the degree of user participation there are three types: explicit crowdsourcing-based RSS collection, implicit crowdsourcing-based RSS collection\awn{,} and partially-labeled fingerprinting. The first two \awn{require }users to report their locations \awn{by marking them manually }on a digital map \awn{whenever a vector of sampled RSS values is stored or uploaded for RM generation}. \awn{The crowdsourced }RSS readings are used directly \awn{with the first method while they are} filtered\awn{ or }combined with other resources \awn{according to the second method}. \awn{The third method }involves less active participation of the users \awn{who only need to agree that RSS values and location information are shared by their mobile device but do not need to manually identify their location on a map}. For example \cite{Wu2013} proposed an approach report\awn{ing} the RSS values of APs \awn{along with the} sampling position \awn{estimated automatically from the data} of the built-in inertial sensors of \awn{the respective} mobile device. \awn{T}hese crowdsourcing based approaches \awn{are less labor intensive (for the provider of the positioning service) but their} performance \awn{is limited by uncertainties introduced through (i) the application of different devices, (ii) manual position indication by the users, or (iii) location estimation with the built-in inertial sensors}.
	
Another popular approach is constructing the radio map in a fast way \awn{by densifying or} adapting a sparse radio map comprising only few \awn{originally measured }RPs and the\awn{ir} associated RSS values. In \cite{Bernardos2010} and \cite{Atia2013} the authors investigated infrastructure based approaches\awn{which} require deploy\awn{ing} specific hardware for RSS monitoring\awn{. The RM is} then construct\awn{ed} or update\awn{d} to \awn{match }the RSS values observed at stationary monitoring points. \awn{The} requirement \awn{of extra installations for RSS monitoring} is in contrast to the potential advantage of FWIPS \awn{that} the existing WLAN/WiFi infrastructure \awn{can be used }without need for additional deployment. Non-infrastructure based methods \awn{may be used to }infer the updated radio map via transfer learning algorithms (e.g.\awn{,} compressive sensing, $l_1$-minimization, manifold alignment) using the sparse radio map or \awn{crowd sourced} RSS readings \awn{and} the assumption that nearby positions have more similar RSS readings than those far away \cite{Pan2007, 6866898}. \awn{B}oth approaches \awn{have been} investigated in the literature. \awn{The focus,} however, \awn{was only }on build\awn{ing} the RM during the offline stage\awn{; they} were rarely applied to the online stage for location estimation at the same time \cite{He2016}.
	
In this paper, the authors propose a scheme to employ backpropagation neural networks (BPNNs) to learn the mapping relationship between RP coordinates and RSS vectors for both stages of FWIPS. BPNN\awn{s}, \awn{widely} used \awn{in} machine learning, \awn{were} so far applied to indoor location estimation in optical, RFID, WLAN/WiFi and dead reckoning based IPSs \cite{Statistik2014, Wagner2012, Xu2016, Soltani2013, Edel2015}. In this paper, BPNN is not only applied to indoor localization (BPNN based localization, BPNN-LA), but also to \awn{fast} radio map construction \awn{starting }from a sparse training radio map (TRM) (BPNN based radio map construction, BPNN-RM). Employing BPNN for both stages of FWIPS, especially for the RM construction with low workload, \awn{has hardly been} investigated in the literature \awn{so far}. We investigate herein the performance of the proposed scenario compared to two popular fingerprinting localization algorithms (FLAs), $k$NN and weighted $k$NN (W$k$NN). Additionally, we analyze the impact of various choices of BPNN design \awn{parameters} via numerical simulations and derive proposals regarding these choices.
	
The structure of the remaining paper is as follows: the principles of an FWIPS are described in Section \ref{sec_2}. In Section \ref{sec_3} definitions of \awn{a} BPNN as well as the proposed scenario of employing BPNN to FWIPS are illustrated. An experimental analysis of the performance using the proposed scheme is presented in Section \ref{perf_analy}.

\section{Principles of FWIPS}\label{sec_2}

\begin{figure}[htb]
			\centering
			\includegraphics[width=0.5\columnwidth]{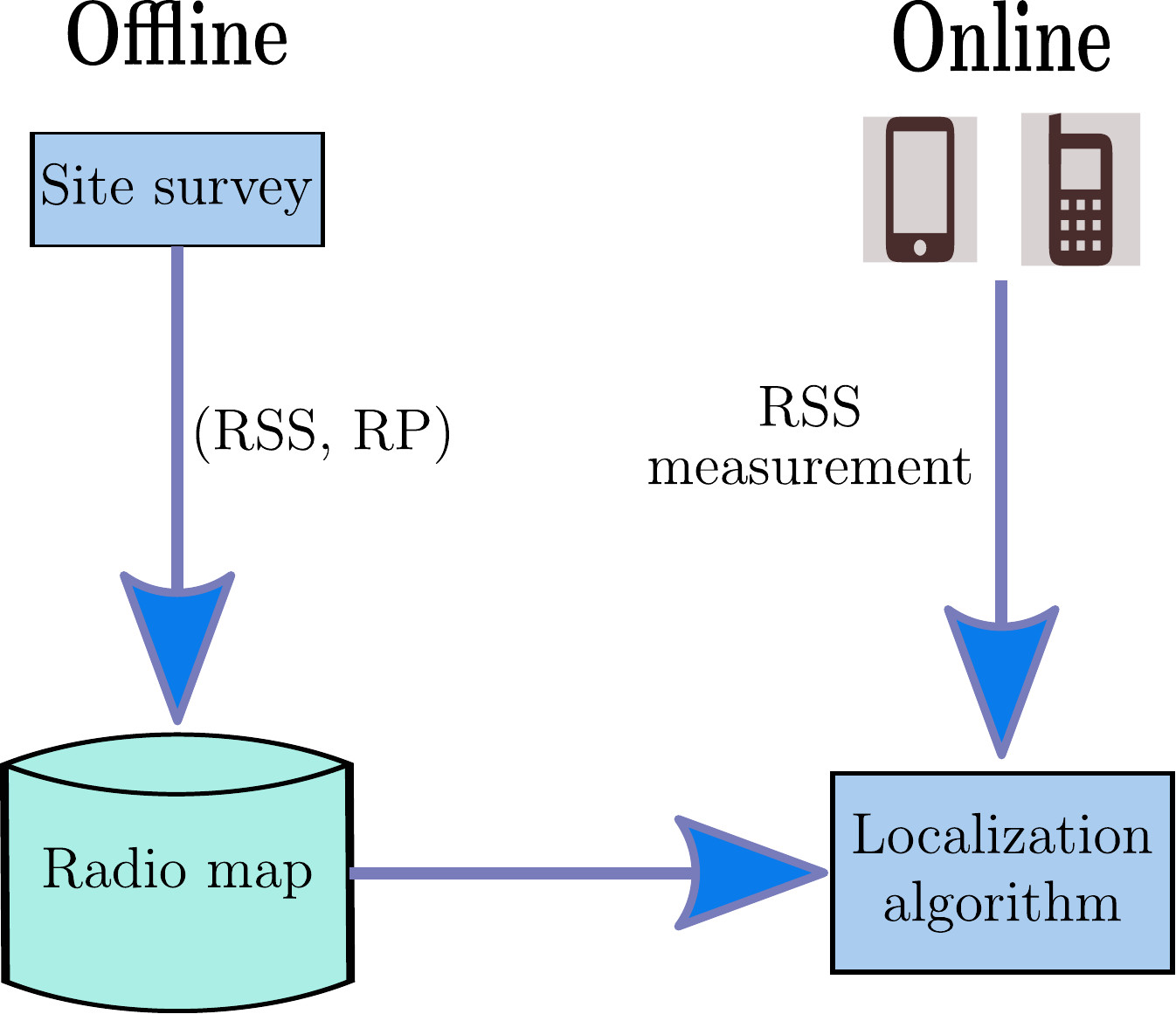}
			\caption{Overview of an FWIPS}
			\label{fig_overview_of_fwips}
		\end{figure}

In this section, the authors give more details on the definitions of an FWIPS, including the deployment of RPs, RSS collection and performance evaluation. A typical FWIPS consists of two stages: offline and online as shown in Fig.\awn{\,}\ref{fig_overview_of_fwips}. During the offline stage, the data required to construct the RM are collected within the RoI covered by WLAN/WiFi signals. The radio map is then employed together with RSS measurements recorded by the user device to estimate the user's location via FLAs within the online stage.
	
\subsection{Offline Stage}\label{sec_2_1}

\begin{figure}[htb]
			\centering
			\includegraphics[width=0.75\columnwidth]{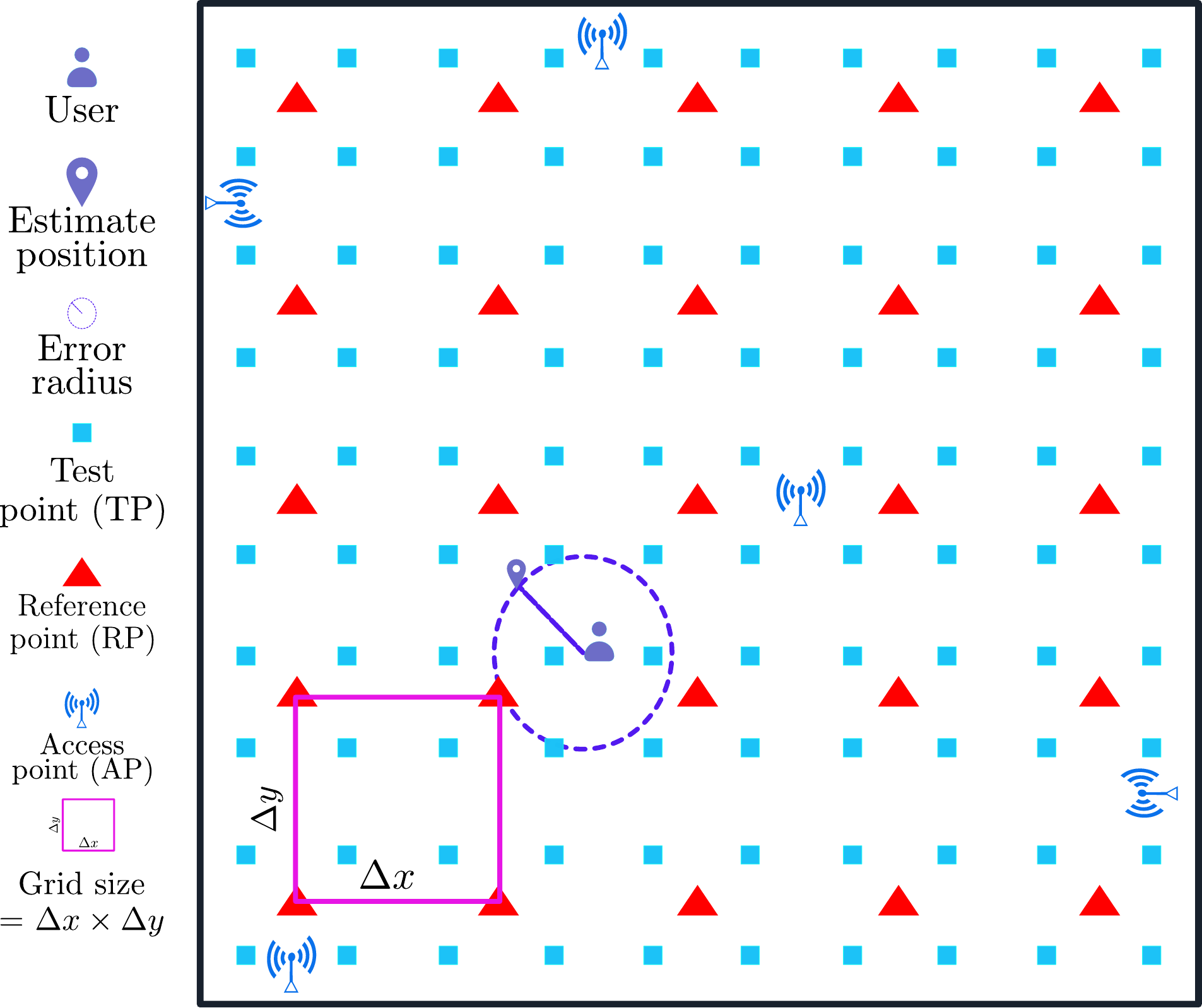}
			\caption{Schematic arrangement of points involved in the deployment, validation and use of an FWIPS}
			\label{fig_grid_deployment_of_fwips}
		\end{figure}

	If the RoI is covered by a sufficient number of APs distributed spatially such that several of them are available if the user device occupies any position within the RoI, no modifications are necessary. Should there be too few APs for positioning, additional APs have to be installed as radio sources for the WIPS. In this paper we assume that the signals of ${N}$ APs can be received within the RoI. For the sake of simplicity we assume herein the RoI is rectangular and the APs are regularly distributed across the RoI as visualized in the schematic map given in Fig.\awn{\,}\ref{fig_grid_deployment_of_fwips}.
	
	${M}$ known locations in the area are selected as RPs. We collect their coordinates in the ${M}\times{D}$ matrix $\mathbf{{R}} = [\mathbf{r}^{(1)},\mathbf{r}^{(2)},\cdots,\mathbf{r}^{({M})}]$ where the $i$-th column $\mathbf{r}^{(i)}\in\mathcal{R}^{{D}}$ is the column-vector of coordinates of the $i$-th RP (e.g.\awn{,} in a 2D scenario $\mathbf{r}^{(i)} = [x^{(i)};y^{(i)}]$). In this paper, we use the grid size ${G} = \Delta x\times\Delta y$ (see Fig.\awn{\,}\ref{fig_grid_deployment_of_fwips}) of the rectangular arrangements of RPs as a measure of the amount of reference data to be provided during the offline stage and thus as a measure of workload and cost. The smaller the grid size, the higher the workload to construct the RM, but the better the anticipated positioning accuracy.  
	
	At all the RPs \awn{the RSS are} sampled \awn{and associated with the respective APs using the data extracted from the beacon frames}. The results are stored in the matrix $\mathbf{{S}} = [\mathbf{s}^{(1)},\mathbf{s}^{(2)},\cdots,\mathbf{s}^{({M})}]$, where each of the ${M}$ columns contains the recorded RSS values of \awn{the }${N}$ APs, i.e.\awn{,} the fingerprint $\mathbf{s}^{(i)} = [RSS^{(i,1)};RSS^{(i,2)};\cdots;RSS^{(i,{N})}]\in\mathcal{R}^{{N}}$ and each column is associated with one RP. If the coordinates of the RPs are not known and these points are not (yet) marked visibly in the physical space, the coordinates need to be determined along with the recording of the RSS measurements. This can be achieved by employing a suitable positioning technology (e.g.\awn{,} a multi-sensor system involving inertial sensors, or a total station). The coordinates are then again assumed as known. The sampling results and the known RP coordinates can then be combined to represent the original radio map (ORM) with the defined grid size.
	
	\subsection{Online Stage}\label{online_stage}
	During the online stage, the ${N}$-dimensional RSS vector $\mathbf{s}^{(t)}\in\mathcal{R}^{{N}}$ is measured at an unknown location $\mathbf{l}^{(t)}\in\mathcal{R}^{{D}}$ by a user who requests the positioning service. The aim is to calculate $\mathbf{l}^{(t)}$ from $\mathbf{s}^{(t)}$ and the RM using an FLA. More details on $k$NN and W$k$NN, two selected FLAs for performance analysis, are presented in \awn{the }following subsections. Herein we will carry out a performance analysis by actually measuring $\mathbf{s}^{(t)}$ at known or independently measured locations such that the error of the positions $\mathbf{l}^{(t)}$ derived from $\mathbf{s}^{(t)}$ can be assessed. We assume that such measurements are actually carried out at ${T}$ locations. We collect the measured RSS in a matrix $\mathbf{S}^{(t)}$ and the corresponding positions in the matrix $\mathbf{R}^{(t)}$. These two matrices represent the validation dataset (VDS).
	
	\subsubsection{$k$NN}
	$k$NN is a method which is widely used in the field of machine learning for classification and clustering. With a selected number $k$ of nearest neighbors $k$NN works in two steps:
	\begin{itemize}
		\item Step 1: find the $k$ nearest neighbors in the RSS space via computing the Euclidean distance between $\mathbf{s}^{(t)}$ and the RSS vectors within the RM. From the view of mathematics the subset $\mathbf{S}^{k\mathrm{NN}}\awn{\subset} \mathbf{S}$ of nearest neighbors is calculated with the condition:
		\begin{eqnarray}
			\centering
			\label{eq_knn_1}
			\begin{split}
			&\mathbf{S}^{k\mathrm{NN}} \awn{\subset} \mathbf{S}:\  
			\mathrm{card}(\mathbf{S}^{k\mathrm{NN}}) =k,\\ 	
			& {\|\mathbf{s}^{(i)} - \mathbf{s}^{(t)}\|}^2_2 \leqslant {{\|\mathbf{s}^{(l)}-\mathbf{s}^{(t)}\|}_2^2}\\
			&\forall \mathbf{s}^{(i)} \in \mathbf{S}^{k\mathrm{NN}}, \mathbf{s}^{(l)} \in \mathbf{S}\backslash \mathbf{S}^{k\mathrm{NN}}
			\end{split}
		\end{eqnarray}
	
		where $\mathrm{card}(\cdot)$ indicates the number of elements of \awn{the} set. \awn{The corresponding $k$ RP locations are collected in the matrix $\mathbf{R}^{k\mathrm{NN}}=
		[\mathbf{r}^{k\mathrm{NN}(1)}, \cdots, \mathbf{r}^{k\mathrm{NN}(k)}]$.}
	
		\item Step 2: estimate the user location $\hat{\mathbf{{l}}}^{(t)}$ as the average of the\awn{se} locations:
		\begin{eqnarray}
			\label{eq_knn_2}
			\hat{\mathbf{{l}}}^{(t)}:=\frac{1}{k}{\sum_{i=1}^{k}{\mathbf{r}^{k\mathrm{NN}(i)}}}
		\end{eqnarray}
\end{itemize}

To evaluate the performance of positioning, the error radius ${e}$, shown in Fig.\awn{\,}\ref{fig_grid_deployment_of_fwips}, is defined as the Euclidean distance between \awn{the }estimated location and the ground truth location of \awn{the }user: 
	\begin{eqnarray}
	\label{eq_error_radius}
	{e^{(t)}}:={\|{\mathbf{l}^{(t)} -\hat{\mathbf{{l}}}^{(t)}}\|_2},\ t=1,2,\cdots, T
	\end{eqnarray}

\awn{For a} statistical analysis  \awn{we will later also use} the mean and standard deviation of \awn{the} error \awn{radii} i.e.\awn{,} $\overline{e}$ and $\sigma_e$ derived from \awn{all }$T$ testing points in the VDS:
	\begin{eqnarray}
	\label{eq_mean_std_error}
	\centering
	\begin{split}
	\overline{e}&:={\sum_{\awn{t}=1}^{T}e^{(\awn{t})}}/{T}\\
	\sigma_e &:=({{\sum_{\awn{t}=1}^{T}(e^{(\awn{t})}-\overline{e})^2}/{(T-1)}})^{1/2}
	\end{split}\
	\end{eqnarray}
	
	\subsubsection{W$k$NN}
	W$k$NN differs from $k$NN only \awn{with respect to (\ref{eq_knn_2}).} Instead of the arithmetic mean W$k$NN uses \awn{a weighted mean with} the respective inverse of the Euclidean distance in the signal space as weight:
	\begin{eqnarray}
	\label{eq_wknn}
	\begin{split}
	\awn{ \hat{\mathbf{l}}^{(t)} }&\awn{:= \left( \sum_{i=1}^{k}w^{(i)}\mathbf{r}^{k\mathrm{NN}(i)}
	\right) / \left( \sum_{i=1}^{k}w^{(i)} \right)
	}
	\end{split}
	\end{eqnarray} 
	where
	\begin{eqnarray}
	\label{eq_wknn_2}
	\begin{split}
	\awn{w^{(i)} }&\awn{:= {1/{\|\mathbf{s}^{k\mathrm{NN}(i)}-\mathbf{s}^{(t)}\|}_2}}
	\end{split}
	\end{eqnarray}
	
	To determine an \awn{appropriate} number $k$ we employ the method typical\awn{ly} used in the field \awn{of} machine learning as given in  \cite{Kulkarni2011}. \awn{Correspondingly, the }upper bound of $k$ is $\lfloor\sqrt{{M}}\rfloor$ (\awn{where }$\lfloor\cdot\rfloor$ returns \awn{the }maximum integer less or equal to $\cdot$). \awn{The concrete choice of $k$ will be discussed later in Section \ref{sec_4_2}.}

\section{BPNN and their application to FWIPS}\label{sec_3}

\awn{An} artificial neural network (ANN) mimic\awn{s} the learning process of \awn{the neurons of }human beings\awn{. Technically} it transfers input data to output \awn{data }\awn{via }interconnected neurons\awn{. The key aspects of ANN design and operation} are (i) the structure in terms of the nodes, layers and activation functions\awn{,} and (ii) the learning algorithm. \awn{We first }present these two concepts \awn{herein}. \awn{Then we discuss the particular training of the ANN causing it to be a BPNN. Finally, we present a }general scenario \awn{for applying BPNNs to WIPS }including RSS sampling, BPNN-LA and BPNN-RM.

\subsection{Design elements of an ANN}\label{sec_3_1}
	\begin{figure}[!h]
		\begin{center}
			\subfigure[A node of \awn{an }ANN]{
				\includegraphics[width=0.4\columnwidth,draft=false]{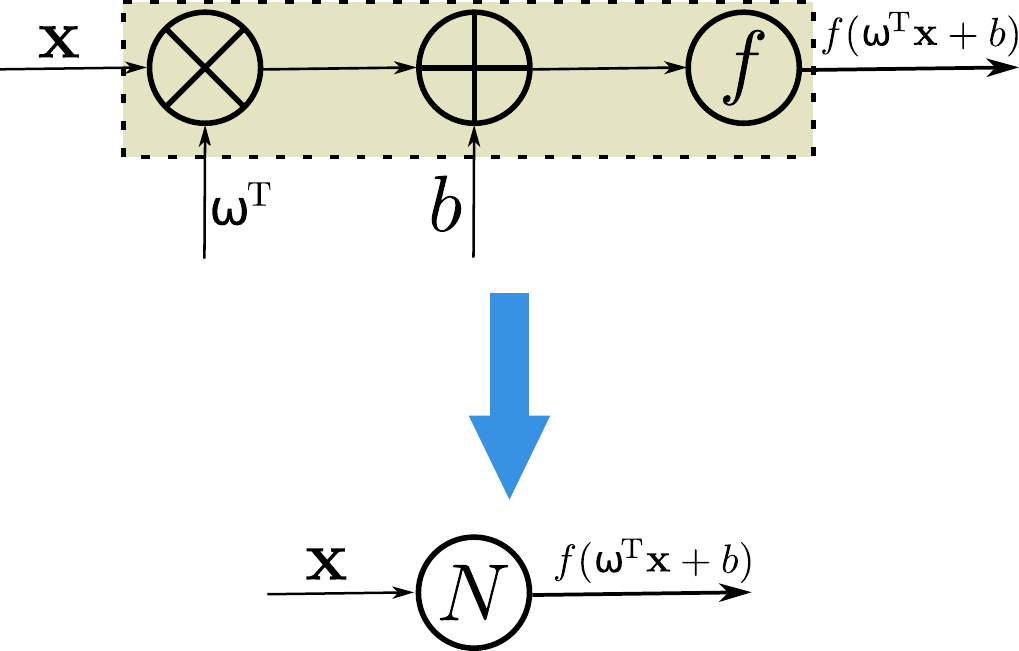}
				\label{fig_single_node}}
			\subfigure[The basic structure of \awn{an }ANN]{
				\includegraphics[width=0.9\columnwidth,draft=false]{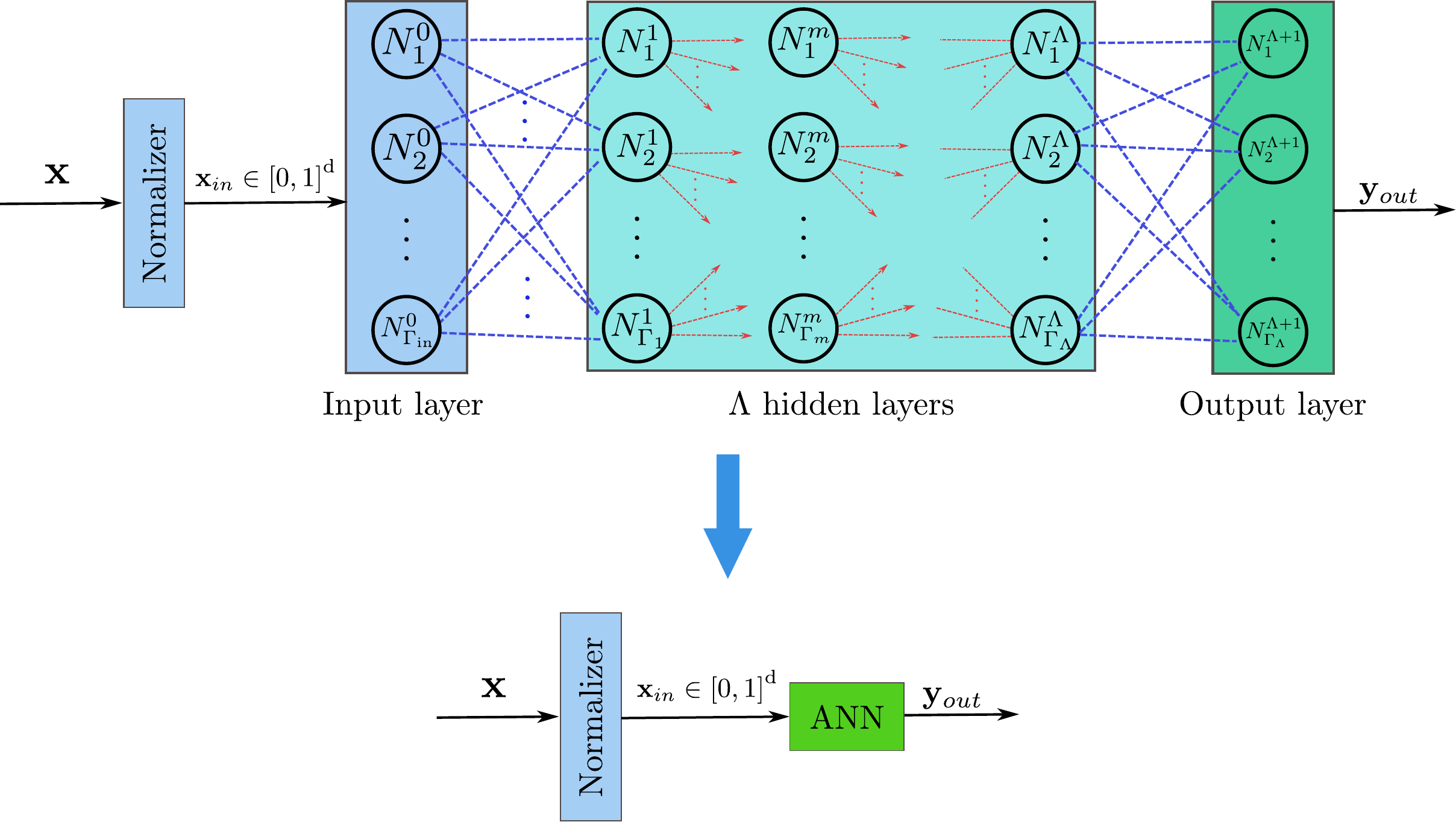}
				\label{fig_ann}}
		\end{center}
		\caption{\awn{Schematic view of a single node and the layered structure of a }BPNN}
		\label{fig_evoluation_view_of_bpnn}
	\end{figure}
	
\subsubsection{Nodes of the ANN}

A node is the elementary unit of \awn{an }ANN. The node works as shown in Fig.\awn{\,}\ref{fig_single_node}. The node takes a column vector $\mathbf{x}$ \awn{(}from the input data or the \awn{preceding }layer\awn{)} as the input, multiplies with the \awn{vector of }weight\awn{s} $\boldsymbol{\omega}$, and adds a \awn{scalar} bias $b$. Th\awn{e} result \awn{of this operations }is used as the argument of a so-called activation function $f$. The evaluation of $f$\awn{, i.e., $y=f(\boldsymbol{\omega}^{\mathrm{T}}\mathbf{x}+b)$,} is the output of the node and \awn{represents---together with the outputs of the other nodes of the same layer---}the input of the \awn{subsequent }layer or the output \awn{of the ANN}. The properties of \awn{each }node are determined by the \awn{activation function, }weight\awn{s} and bias.

\subsubsection{Layers of the ANN}
The nodes are arranged into three types of layers: input layer, hidden layers and output layer. The input layer which has ${\Gamma}_{\mathrm{in}}$ nodes, transforms the general input $\mathbf{x}\in\mathcal{R}^{\mathrm{d}}$ into the space $\mathcal{R}^{\mathrm{\Gamma}_{\mathrm{in}}}$. This dimension transformation depends on the specific applications and the design of \awn{the }ANN. There is no dimension transformation in the input layer in this paper. With a given activation function (e.g.\awn{,} sigmoid \awn{or }linear) of the input layer, the input domain of $f$ is often limited, so that we apply a normalizer to transform any range of the input vector \awn{components}to the domain of $f$. \awn{With the activation functions chosen herein this domain is the interval $[0,1]$ such that the} normalizer maps \awn{each component of }$\mathbf{x}$ into \awn{that interval} via \awn{an affine }transformation \awn{consisting of a }scaling $\mathbf{S}_{\mathrm{in}}\in\mathcal{R}^{\mathrm{d}\times\mathrm{d}}$ and \awn{translation }$\mathbf{h}_{\mathrm{in}}\in\mathcal{R}^{\mathrm{d}}$\awn{.} \awn{The elements of the diagonal matrix $\mathbf{S}_{\mathrm{in}}$} and of $\mathbf{h}_{\mathrm{in}}$ are determined by the range of input data. The \awn{input }$\mathbf{x}_{\mathrm{in}}$ \awn{to the first hidden layer of the ANN} is \awn{thus }calculated by:
	\begin{eqnarray}
	\label{eq_x_in}
	\begin{split}
		\mathbf{x}_{\mathrm{in}}=\mathbf{S}_{\mathrm{in}}\cdot\mathbf{x}+\mathbf{h}_{\mathrm{in}}
	\end{split}
	\end{eqnarray} 
	
\awn{All nodes of a specific layer }share the same activation function but have different weights and biases. Denoting the weights and biases of the input layer as $\boldsymbol{\Omega}_{\mathrm{in}}=[\boldsymbol{\omega}_{1}^{0},\cdots,\boldsymbol{\omega}_{{\Gamma}_{\mathrm{in}}}^{0}]$ and $\mathbf{b}_{\mathrm{in}}=[b_{1}^{0},\cdots,b_{{\Gamma}_{\mathrm{in}}}^{0}]$\awn{,} respectively, the output, $\mathbf{f}_{\mathrm{in}}(\boldsymbol{\Omega}_{\mathrm{in}}^{\mathrm{T}}\mathbf{x}+\mathbf{b}_{\mathrm{in}})\in\mathcal{R}^{{\Gamma}_{\mathrm{in}}}$ of \awn{the }input layer is the input \awn{to each node of }the first hidden layer.
	
The \awn{required }number of hidden layers depends on the application, especially on the non-linearity of the relation between input and output \cite{Larochelle2009}. Generally, there are $\Lambda$ hidden layers and a different number of nodes in each hidden layer. From the training and convergence perspective, \awn{$\Lambda$} should not be too big, especially in the application to FWIPS \cite{Larochelle2009}. Usually there is just one hidden layer \cite{Brunato2005}. We will later \awn{analyze} the performance with up to 3 hidden layers whose respective number of nodes is up to 30 for each of them.
	
As for the output layer, the number of nodes equals the dimension of \awn{the output}. Except the input layer, there are $\Lambda+1$ layers in total. Here we denote the activation function, the weights and the biases of \awn{the }$m^{th}$ \awn{layer }($m = 1,\cdots,\Lambda,\Lambda+1$) as $\mathbf{f}_{m}$, $\boldsymbol{\Omega}_{\mathrm{m}}=[\boldsymbol{\omega}_{1}^{m},\cdots,\boldsymbol{\omega}_{{\Gamma}_{\mathrm{m}}}^{m}]$, $\mathbf{b}_{\mathrm{m}}=[b_{1}^{m},\cdots,b_{{\Gamma}_{\mathrm{m}}}^{m}]$\awn{,} respectively. In the cases of positioning and of radio map construction, $\Gamma_{\Lambda+1}$ equals the dimension of the coordinates and the number of available APs\awn{,} respectively i.e.\awn{,} $D$ and $N$ in this paper. The basic structure of the ANN is presented in Fig.\awn{\,}\ref{fig_ann}. The design parameters that influence the performance of the ANN are the type of activation function, the number of hidden layers ($\Lambda$), the numbers of nodes in the hidden layers ($\boldsymbol{\Gamma}$), the weights (${\boldsymbol{\Omega}}$) and the biases ($\mathbf{B}$) for the nodes. In this paper, we use the sigmoid function and a linear function as the activation functions for the hidden layers and the output layer\awn{,} respectively. Formally, the output $\mathbf{y_{out}}$ as shown in Fig.\awn{\,}\ref{fig_ann} is:
	\begin{eqnarray}
	\label{eq_ann_out}
	\begin{split}
	\mathbf{y}_{out} = & \mathbf{f}_{\Lambda+1}(
	{\boldsymbol{\Omega}}^{\mathrm{T}}_{\Lambda+1}
	\mathbf{f}_\Lambda(
	{\boldsymbol{\Omega}}^{\mathrm{T}}_{\Lambda}
	\mathbf{f}_{\Lambda-1}(
	{\boldsymbol{\Omega}}^{\mathrm{T}}_{\Lambda-1}
	\cdots\mathbf{f}_1(
	{\boldsymbol{\Omega}}^{\mathrm{T}}_{1}
	\mathbf{f}_{in}(
	{\boldsymbol{\Omega}}^{\mathrm{T}}_{in}\mathbf{x}\\
	&+\mathbf{b}_{in})
	+\mathbf{b}_1)
	+\cdots+ \mathbf{b}_{\Lambda-1})
	+\mathbf{b}_\Lambda)
	+\mathbf{b}_{\Lambda+1})
	\end{split}
	\end{eqnarray}
	\subsubsection{Training of the ANN}\label{training_ANN}
	The purpose of the training is to determine the weights and biases such that the error $\delta \mathbf{y} = \mathbf{y}_{target} -\mathbf{y}_{out}$ is minimized using the training data set $\{\mathbf{x}_{\mathrm{training}},\mathbf{y}_{\mathrm{target}}\}$ while the activation functions, number of hidden layers and number\awn{s} of nodes within each layer are fixed. Backward error propagation \cite{Hagan1995} is an established approach to efficiently carry out this optimization. As for the implementation of training the ANN, a given \awn{radio map} will be divided \awn{arbitrarily }into three \awn{data}sets: training, validation and testing \awn{data}set. The training \awn{data}set is used to update the weights and biases\awn{,} the validation set is employed to check the mean square error (MSE) of the \awn{output} with the updated weights and biases\awn{, and the testing data set is used for quality control after completion of the training}. 
	
	The training process stops \awn{when certain conditions are fulfilled. In the ANN implementation used herein three conditions }\awn{are checked }as shown in Table \ref{training_conditions}, \awn{and training stops if any of them is fulfilled:} (i) \awn{the MSE calculated from the validation dataset }is no more than the maxim\awn{um admissible} error\awn{;} (ii) the number of \awn{training }epoch\awn{s}\awn{\footnote{Here we use the term 'epoch' instead of 'iteration' to indicate the training steps because each step typically includes a batch of training points, and each training point requires one iteration.}} reached the maxim\awn{um admissible} number of epoch\awn{s;} (iii) the MSE \awn{calculated from the validation dataset} increases continuously \awn{over more than} the maxim\awn{um admissible} number of \awn{epochs with failed} validation. \awn{The weights and biases as of the stopping epoch are selected as training result for the first two cases, and those of the epoch at which the MSE starts increasing for the third case}. 
	
	After stopping the training process, the testing \awn{data}set is applied\awn{.} If the MSE of the \awn{resulting output} is comparable \awn{to the one from the validation dataset} the training process of \awn{the }ANN \awn{is }finished. \awn{O}therwise the trained ANN will be treated as unreasonable because of the inconsistency of the MSE between validation and testing \awn{data}set. This inconsistency is caused by \awn{an inappropriate }division of the \awn{available data into the three datasets}. \awn{In this case }the training of \awn{the} ANN will be \awn{repeated with a different partitioning of the data} until the MSE of the trained ANN is consist\awn{ent} \awn{for }both validation and testing \awn{data}set. \awn{The }initial value\awn{s} of \awn{the }weights and biases for all \awn{neurons }are initialized randomly \awn{when starting the training. This is }the established method \cite{Hagan1995}. We will later evaluate the influence of this random initialization on the performance of \awn{the }ANN in \awn{S}ection \ref{perf_analy}. 
	\begin{table}[!htb]
		\centering
		\caption{Training conditions}
		\label{training_conditions}
		\begin{tabular}{lcc}
			\hline
			&BPNN-LA&BPNN-RM\\
			\hline
			Max. \#\textsuperscript{*} epochs & 1000 & 1000\\
			Max. error & 0.25\awn{\,}$\mathrm{m}^2$ & 1\awn{\,}$\mathrm{dB}^2$\\
			Max. \# failed validation\awn{s} & 6 & 6\\
			\hline
			{\scriptsize\textsuperscript{*} \#: number of}
		\end{tabular}
	\end{table} 
	
\subsection{Chain rule for gradient descent optimization of BPNN}\label{sec_3_2}

To compute the optimal weights and biases of the nodes, gradient descent is applied \awn{for minimizing} the squared training error ${\hat{F}}(y) = {\delta \mathbf{y}}^{\mathrm{T}}{\delta \mathbf{y}}$. This optimization is carried out iteratively. The weights and biases of epoch \awn{$t+1$ }depend \awn{on }those of the previous epoch and \awn{on }the gradient descent\awn{:}
	\begin{eqnarray}
	\label{eq_chainrule_1}
	\begin{split}
	\boldsymbol{\Omega}_{m}^{i,j}(t+1)&=\boldsymbol{\Omega}_{m}^{i,j}(t)-\eta\cdot{g_m(\boldsymbol{\Omega}_{m}^{i,j})},\ \boldsymbol{\Omega}_{m}\in\mathcal{R}^{\Gamma_{\mathrm{m-1}}\times\Gamma_{\mathrm{m}}}\\
	\mathbf{b}_{m}^{\awn{j}}(t+1) &= \mathbf{b}_{m}^{\awn{j}}(t) -\eta\cdot{g_m(\mathbf{b}_{m}^{\awn{j}})},\ \mathbf{b}_{m}\in\mathcal{R}^{\Gamma_{\mathrm{m}}}
	\end{split} 
	\end{eqnarray}
where \awn{$j$ is the index of the node, $i$ the index of the weight per node, }$\eta$ the learning rate, $\Gamma_0 := \Gamma_{\mathrm{in}}$ and $g_m(\cdot)$ \awn{are }the gradient\awn{s} of ${\hat{F}}(y)$ in the $\mathbf{\Omega}$-space or $\mathbf{b}$-space of the $m^{th}$ layer\awn{, respectively}. Assuming that the inputs to the activation functions of the $m^{th}$ layer and their outputs are $\mathbf{y}_{in}^{m}$ and $\mathbf{y}_{out}^{m}$ respectively, we have:
	\begin{eqnarray}
	\label{eq_chainrule_2}
	\begin{split}
	\mathbf{y}_{in}^{m} &= {\boldsymbol{\Omega}_{m}}^{\mathrm{T}}\mathbf{y}_{out}^{m-1}+\mathbf{b}_{m}\\
	\mathbf{y}_{out}^{m} & = \mathbf{f}_{m}(\mathbf{y}_{in}^{m})
	\end{split}
	\end{eqnarray}
Therefore, \awn{according to the chain rule }the gradient\awn{s} w.r.t. $\boldsymbol{\Omega}_{m}^{i,j}$ and $\mathbf{b}_{m}^{k}$ are:
	\begin{eqnarray}
	\label{eq_chainrule_3}
	\begin{split}
	g_m(\boldsymbol{\Omega}_{m}^{i,j})& =({\partial{{\hat{F}}(y)}}/{\partial{\mathbf{y}_{in}^{m}}})\cdot({\partial{\mathbf{y}_{in}^{m}}}/{\partial{\boldsymbol{\Omega}_{m}^{i,j}}})\\
	g_m(\mathbf{b}_{m}^{k}) &= ({\partial{{\hat{F}}(y)}}/{\partial{\mathbf{y}_{in}^{m}}})\cdot({\partial{\mathbf{y}_{in}^{m}}}/{\partial{\mathbf{b}_{m}^{k}}})
	\end{split}
	\end{eqnarray} 
	The first term \awn{on the right side of }(\ref{eq_chainrule_3}) is redefined as $\mathbf{s}_m := \partial{{\hat{F}}(y)}/\partial{\mathbf{y}_{in}^{m}}$. The weights and biases of the $m^{th}$ layer are then updated \awn{according to}: 
	\begin{eqnarray}
	\label{eq_chainrule_4}
	\begin{split}
	\boldsymbol{\Omega}_{m}(t+1)&=\boldsymbol{\Omega}_{m}(t)-\eta\cdot{(\mathbf{s}_m\awn{\cdot}(\mathbf{y}_{\mathrm{out}}^{m-1})^{\mathrm{T}})^{\mathrm{T}}}\\
	\mathbf{b}_{m}(t+1) &= \mathbf{b}_{m}(t) -\eta\cdot{\mathbf{s}_m}
	\end{split} 
	\end{eqnarray}
	According to the chain rule $\mathbf{s}_m$ can be calculated from the corresponding vector $\mathbf{s}_{m+1}$ of the \awn{subsequent }layer:
	\begin{eqnarray}
	\label{eq_chainrule_5}
	\begin{split}
	\mathbf{s}_m & = ({\partial{(\mathbf{y}_{in}^{m+1})}}/{\partial{(\mathbf{y}_{in}^{m})^{\mathrm{T}}}})\cdot({\partial{{\hat{F}}(y)}}/{\partial{(\mathbf{y}_{in}^{m+1})^{\mathrm{T}}}})\\
	&=\dot{\digamma}(\mathbf{y}_{in}^m)(\boldsymbol{\Omega}_{m+1})^{\mathrm{T}}\mathbf{s}_{m+1}
	\end{split}
	\end{eqnarray}
	where $\dot{\digamma}(\mathbf{y}_{in}^m)$ is a diagonal matrix of derivatives of the \awn{activation }functions \awn{with respect to $\mathbf{y}_{in}^m$}.
	\begin{eqnarray}
	\label{eq_chainrule_6}
	\dot{\digamma}(\awn{\mathbf{y}_{in}^m}) = 
	\left[
	\begin{array}{cccc}
	\dot{f}_m({\mathbf{y}_{in}}^m(1)) & \cdots & 0\\
	\vdots & \vdots & \vdots\\
	0 & \cdots & \dot{f}_m(\mathbf{y}_{in}^m(\Gamma_{m}))
	\end{array}
	\right]
	\end{eqnarray}
	In this way, gradient descent learning works \awn{by }backpropagation: $\mathbf{s}_{\Lambda+1}\rightarrow \mathbf{s}_\Lambda \rightarrow \cdots\rightarrow \mathbf{s}_1$. To sum up, each epoch of BPNN training works via forward and backward propagation according to:
	\begin{eqnarray}
	\label{eq_chainrule_7}
	\begin{split}
	\mathbf{y}_{out}^{0} := & \mathbf{x}\\
	\mathbf{y}_{out}^{m+1} =& \mathbf{f}_{m+1}(\boldsymbol{\Omega}_{m+1}^{\mathrm{T}}\mathbf{y}_{out}^{m} + \mathbf{b}_m),\ m = 0,1,\cdots\, \Lambda \\
	\mathbf{y}_{out} = & \mathbf{y}_{out}^{\Lambda+1}\\
	\mathbf{s}_{\Lambda+1} = & -2\dot{\digamma}(\mathbf{y}_{in}^{\Lambda+1})\delta \mathbf{y}\\
	\mathbf{s}_{n} = & \dot{\digamma}(\mathbf{y}_{in}^n)(\boldsymbol{\Omega}_{n+1})^{\mathrm{T}}\mathbf{s}_{n+1},\ n = \Lambda, \Lambda-1, \cdots, 1
	\end{split}
	\end{eqnarray}
	\subsection{BPNN based radio map construction \& localization }\label{sec_3_3}
\begin{figure*}[!t]
		\centering
		\includegraphics[width=1.4\columnwidth]{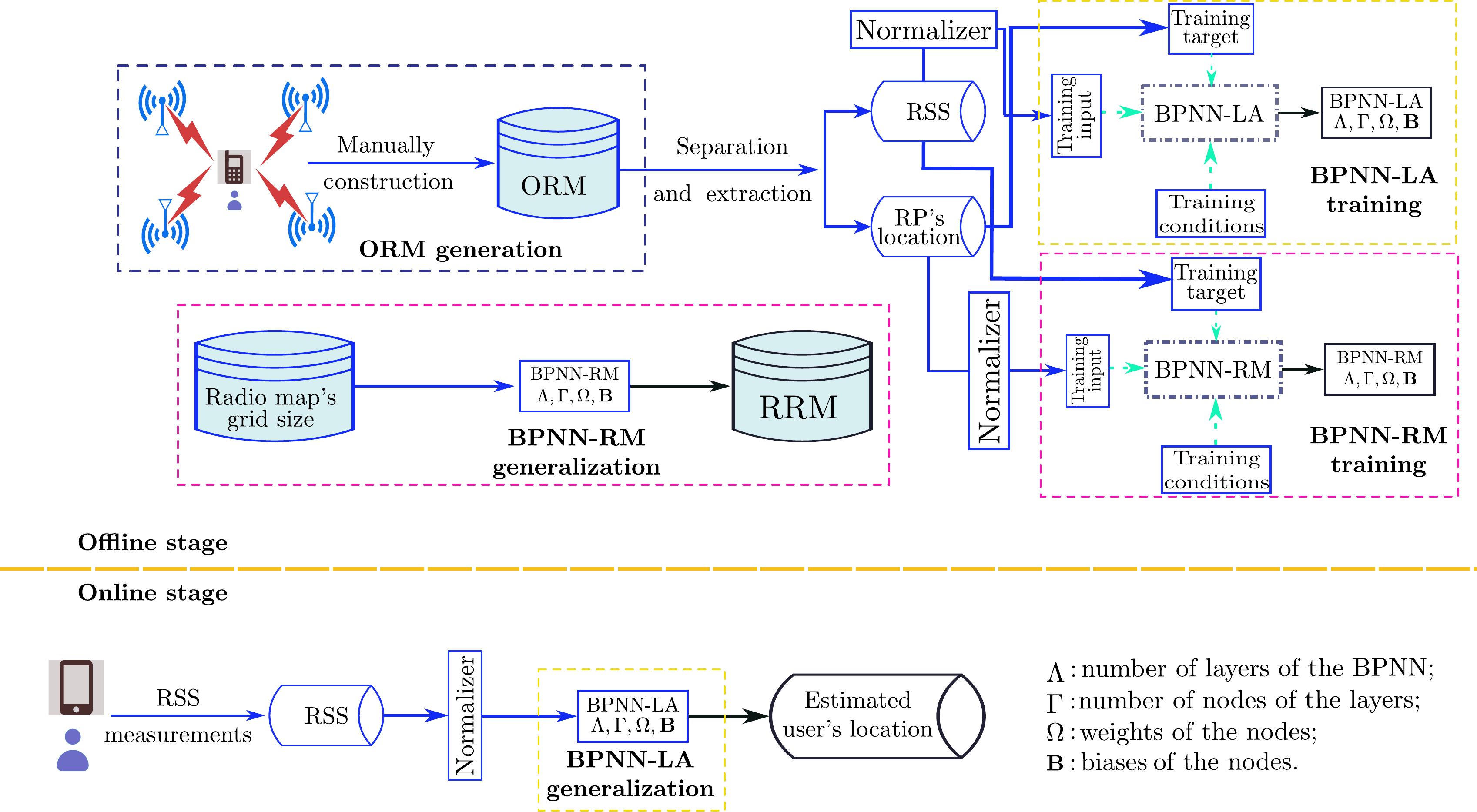}
		\caption{Systematic view of the proposed algorithms}
		\label{fig_systematic_flowchart}
	\end{figure*}

	On the basis of BPNN we propose an algorithm for radio map construction and indoor localization. The systematic view of the proposed approach is presented in Fig.\awn{\,}\ref{fig_systematic_flowchart}.
	
\subsubsection{ORM generation module}

At given RPs a surveyor uses the sampling device (e.g. a mobile phone) to collect the RSS from all available APs within the RoI. In this process, the grid size is relatively large to keep the workload low. The coordinates of the RPs and the corresponding RSS vectors are stored in a table which represents the ORM. In order to mitigate the measurement noise, the measurements can be filtered before storing the discrete representation of the spatially continuous signal strength fields as ORM. In the later experiments we will only reduce the impact of noise by averaging multiple RSS measurements taken at each RP within a short time interval.

\subsubsection{BPNN-RM training \& generalization module}
	
This module consists of two parts: training of BPNN and RM generation using the trained BPNN. The module is evoked if (i) a denser discrete representation of the RM is required than the one available as ORM or (ii) if a continuous representation of the RM is required such that the (expected) signal strength of any AP \awn{and---if need be---also of the corresponding spatial derivatives} can be calculated for any location within the RoI. \awn{We denote such an RM, derived using the BPNN, as reconstructed radio map (RRM) subsequently. }{The coordinates of the $M$ RPs stored as part of the ORM are normalized} and then used as the input data for the estimation of optimum weights and biases according to the algorithm described in the previous section. The signal strengthens corresponding to the above inputs in the ORM are the training targets for this BPNN. The BPNN-RM is trained according to the process presented in Section \ref{training_ANN}. 
	
The RRM generation is the process \awn{of generalizing } the trained BPNN. Given a specific \awn{desired} grid size of \awn{the }RRM a set of \awn{corresponding} coordinates within the RoI is generated and normalized. The normalized coordinates {are} the input to the trained BPNN-RM. Combining the \awn{output} vectors with the \awn{above input }coordinates \awn{yields} the RRM.

\subsubsection{BPNN-LA training \& generalization module}
	This module includes two parts: training and generalization (i.e. applying the trained BPNN to localization). During the training, the normalizer transforms the RSS to $[0,1]$ since the activation functions of the hidden layers \awn{are sigmoid }in this paper. The normalized RSS vectors and the corresponding RP locations are the training input and training target respectively. \awn{The }constraints shown in Table \ref{training_conditions} \awn{are used as criteria during the training}. With a given training dataset, the training of BPNN-LA follows the procedures in Section \ref{training_ANN}. The trained BPNN ($\Lambda, \boldsymbol{\Gamma}, \boldsymbol{\Omega}, \mathbf{B}$) is saved for the generalization within the online stage.

\section{Experimental Performance Analysis}\label{perf_analy}
\subsection{Testbed}\label{sec_4_1}	
\begin{figure}[!htb]
		\centering
		\includegraphics[width=0.55\columnwidth]{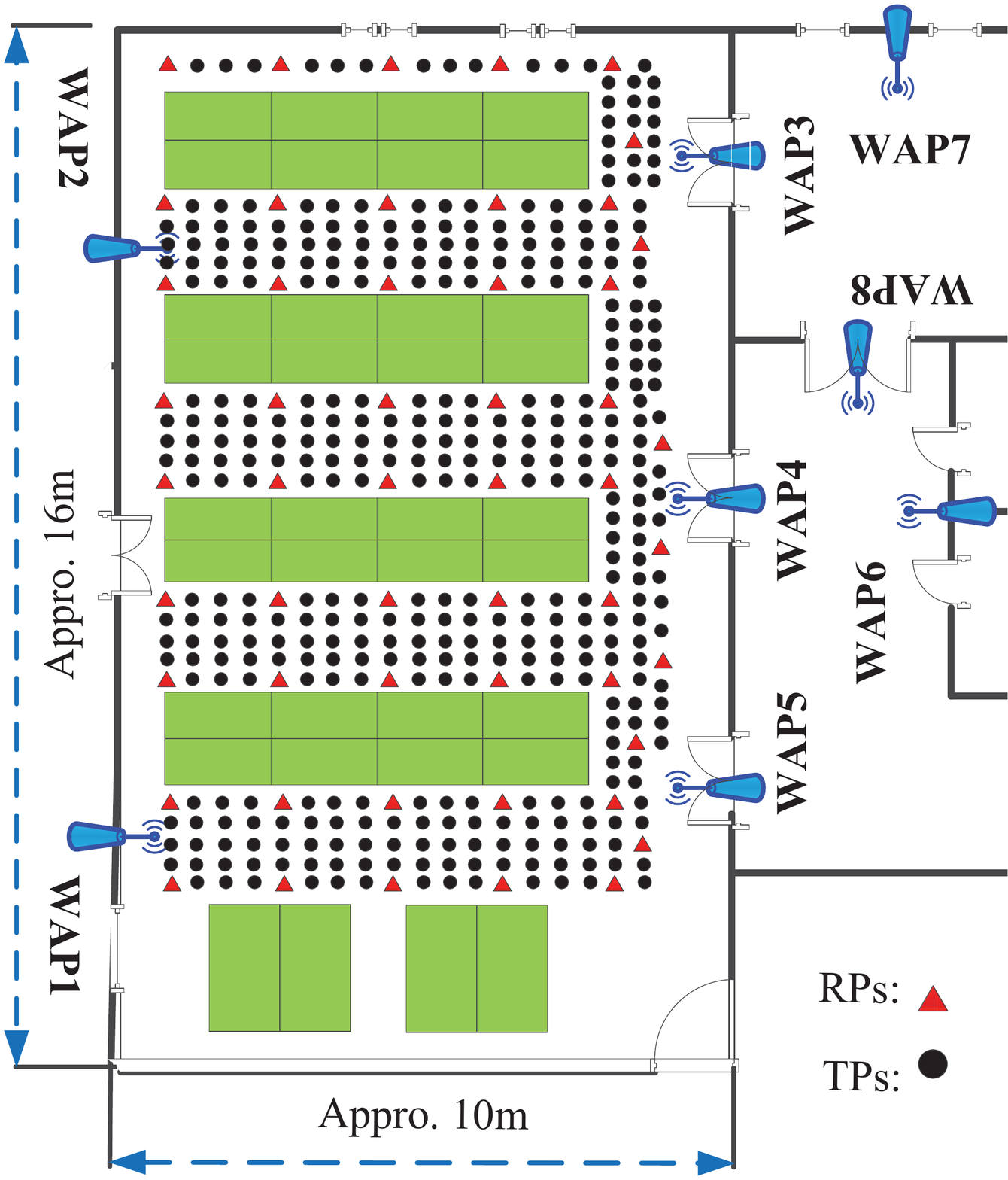}
		\caption{Indication of Testbed}
		\label{fig_testbed_settings}
	\end{figure}

In this section, we test our proposed approach using real measurements from the $10^{th}$ floor of a building at Harbin Institute of Technology, depicted in Fig.\awn{\,}\ref{fig_testbed_settings}\footnote{The dataset was created while the first author was with the Communication Research Center, Harbin Institute of Technology, Harbin, P.R. China as master student.}. There are 8 APs in the experimental area, which are attached stably to the wall at a height of 2\awn{\,}$\mathrm{m}$ from the floor. \awn{RSS were recorded at points arranged in a regular grid of $0.5\times0.5\,{\mathrm m}^2$}\awn{yielding} an ORM with \awn{a }grid size of 0.25\awn{\,}$\mathrm{m}^2$. It is subsequently annotated as $\mathrm{ORM_{0.25}}$. \awn{Some of these points were later used as RPs for positioning or RM generation, others for testing only. For the former purpose a training radio map (TRM)} with larger grid size \awn{was then }obtained by down-sampling from \awn{the }$\mathrm{ORM_{0.25}}$. The sampling and preprocessing of the RSS values are described in \cite{69}. 

\subsection{BPNN based indoor localization}\label{sec_4_2}

In this \awn{section an }experimental analysis \awn{of the quality of localization using BPNN and of} the \awn{related design }parameters \awn{is }presented. All the following simulations are carried out \awn{using }MATLAB R2015a on Euler, a high performance computing cluster \awn{of }ETH. First, an example is given to show how to determine the locally optimal number of layers and nodes \awn{with }a given TRM \awn{(grid size)}. Then a table shows all the locally optimal parameters w.r.t. the mean error radius for various \awn{TRM }grid sizes as well as \awn{for }different number\awn{s} of hidden layers. A detailed analysis \awn{of }parameter selection, computational complexity and cumulative positioning {error }is presented afterwards. Furthermore, since the weights and biases of \awn{the }BPNN are initialized randomly, \awn{the }simulation\awn{s} are carried out \awn{100 times using} the same \awn{design }parameters in order to \awn{also take }the influence of the random initialization \awn{into account}. \awn{For this purpose }we collect the mean and standard deviation \awn{resulting from each of the }100 simulations in the vector\awn{s} $\overline{\mathbf{e}}=[\overline{e}_1,\overline{e}_2,\cdots,\overline{e}_{100}]$ and $\boldsymbol{\sigma}_e =[\sigma_{e_1},\sigma_{e_2},\cdots,\sigma_{e_{100}}]$\awn{. We }define the uncertainty \awn{due to }random initialization as the standard deviation of $\overline{\mathbf{e}}$ and $\boldsymbol{\sigma}_e$:
	\begin{eqnarray}
	\centering
	\label{uncertainty}
	\begin{split}
		\sigma_{\overline{\mathbf{e}}} &:= ({\sum_{i=1}^{100}
                              (\overline{\mathbf{e}}(i)-\awn{\overline{\overline{\mathbf{e}}}}
			)^2/99})^{1/2}\\
	\sigma_{\boldsymbol{\sigma}_e} &:=(\sum_{i=1}^{100}(
 		\boldsymbol{\sigma}_e(i)-
		\awn{\overline{\boldsymbol{\sigma}_e}}
            )^2/99)^{1/2}
	\end{split}
	\end{eqnarray}
	where $\overline{\,\cdot\,}$ returns the mean value of $\cdot$. As for the number \awn{$k$ of }nearest neighbors \awn{for }$k$NN and W$k$NN, according to the rule cited in Section \ref{online_stage} we  select it \awn{according }to the number \awn{of }RPs in the TRM. In this paper, there are 139 RPs \awn{within the ORM}. Therefore, the maximal value of $k$ \awn{should be} 11 with the grid size of 0.25\awn{\,}$\mathrm{m}^2$ and 3 in the \awn{case of a }grid size {of }9\awn{\,}$\mathrm{m}^2$.
	\subsubsection{Locally optimal parameters of $\mathrm{TRM}_{0.25}$}
	To present how we determine the locally optimal parameters $\Lambda$ and $\boldsymbol{\Gamma}$ we take $\mathrm{TRM}_{0.25}$ (i.e\awn{., }$\mathrm{ORM}_{0.25}$) as an example. With a given TRM, a BPNN with the specific number of hidden layers and neurons is employed to learn the mapping between the RSS vectors and the corresponding RP locations. The trained BPNN is then generalized to the VDS for performance evaluation. The parameters which achieve the minimal mean error radius (MER) are treated as the locally optimal ones according to Section \ref{sec_3_3}.  
		\begin{figure}[htb]
			\begin{center}
				\subfigure[Mean error of BPNN-LA with 1 hidden layer (HL1)]{
					\includegraphics[width=0.75\columnwidth,draft=false]{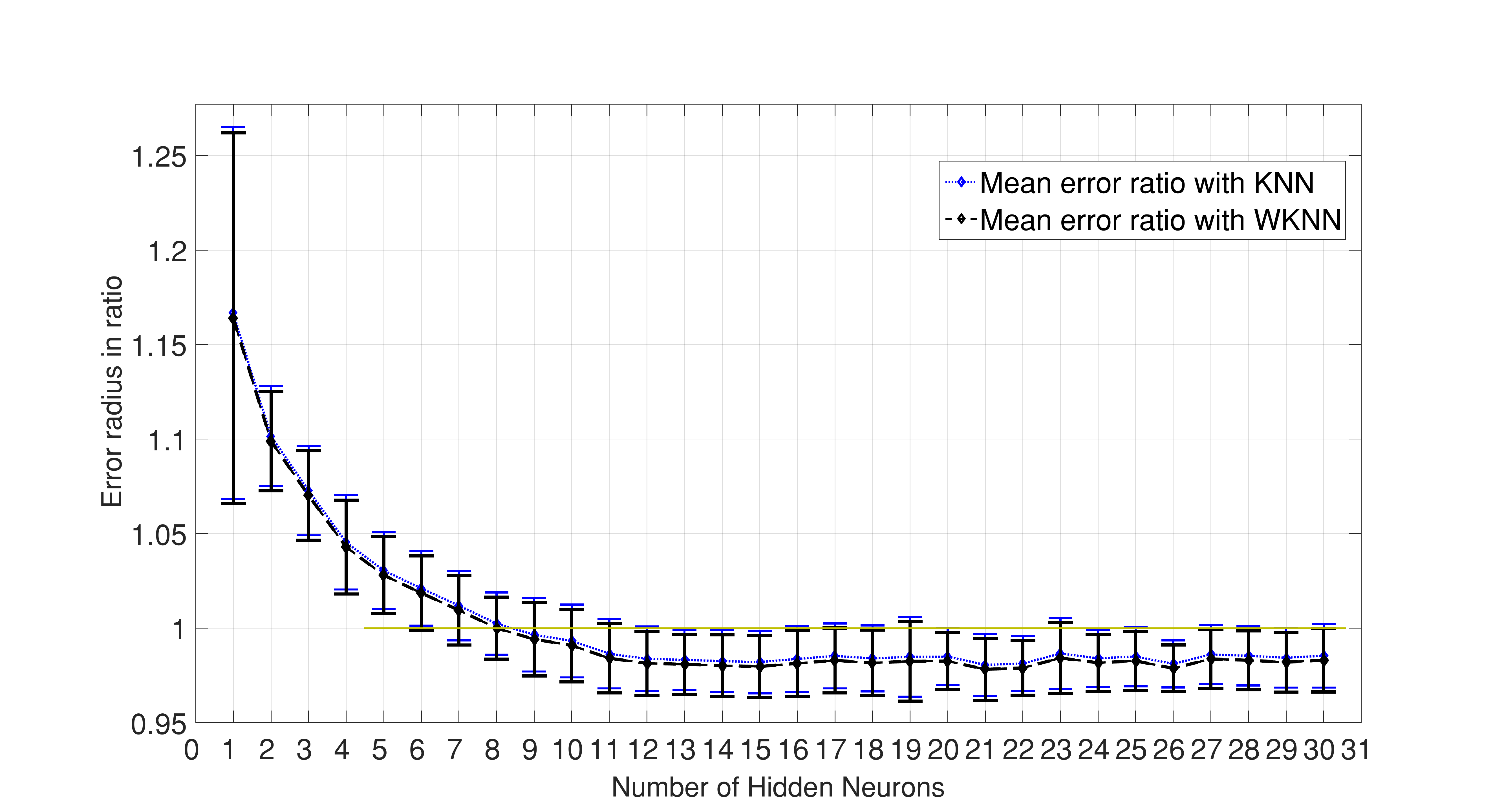}
					\label{fig_bpnnla_mean_error}}
				\subfigure[Ratio of MER of BPNN-LA with 2 hidden layers (HL2)]{
					\includegraphics[width=0.6\columnwidth,draft=false]{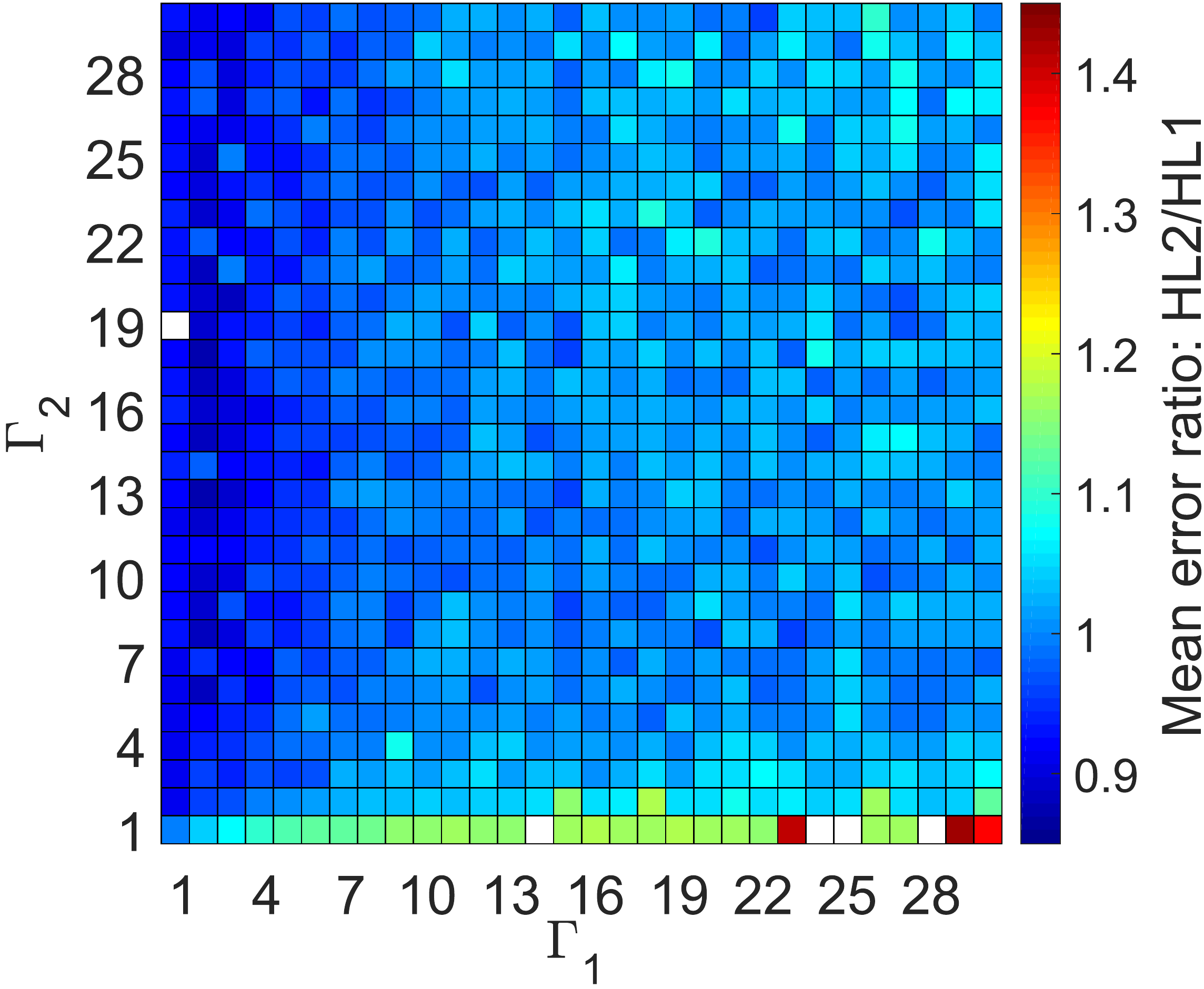}
					\label{fig_bpnnla_hl2_all}}
			\end{center}
			\caption{Error of BPNN-LA with $\mathrm{TRM}_{0.25}$}
			\label{fig_error_of_bpnnla_0_25}
		\end{figure}
	
	\awn{In }Fig.\awn{\,}\ref{fig_bpnnla_mean_error} we \awn{show} the ratio of MER between BPNN-LA and $k$NN as well as W$k$NN \awn{for different numbers }of neurons\awn{. Since }$k$NN and W$k$NN \awn{do not have  neurons, }the variation of ratio in the figure is \awn{exclusively} due to the variation of quality of \awn{the }BPNN solution. A ratio\awn{$<$1} indicates \awn{that the BPNN solution is more accurate than the }$k$NN or W$k$NN \awn{one}. As shown in Fig.\awn{\,}\ref{fig_bpnnla_mean_error}, the MER decreases fast as the number of neurons increases from 1 to 11 for a BPNN with 1 hidden layer. \awn{It hardly changes any more if the number of neurons is further increased.} The locally optimal number of neurons \awn{in this case of} 1 hidden layer (HL1) i.e.\awn{,} the one yielding the minimal MER is {21} in this example. However, taking the uncertainty of the MER into \awn{account}, the plot shows that the MER is stable for $\awn{\Gamma_1}\geq{9}$. Also, for $\awn{\Gamma_1}\geq{9}$, the MER of BPNN-LA is slightly smaller than \awn{the one} obtained using $k$NN and W$k$NN. The standard deviation of \awn{the }error radius is almost independent of the number of neurons if $\awn{\Gamma_1}\geq{2}$. 
	
	With the same process, we can obtain the locally optimal parameters for multiple hidden layers (MHLs). In Fig.\awn{\,}\ref{fig_bpnnla_hl2_all} we present the ratio of MER between BPNN-LA with 2 hidden layers (HL2) and 1 hidden layer layer (HL1) depending on the number of neurons of the layer\awn{s (white color indicates that the value is larger than 1.5)}. It is shown that the MER is larger for HL2 than for HL1 for most \awn{combinations} of numbers of neurons. Furthermore, we see that the MER of HL2 is almost independent of $\awn{\Gamma_2}$, the number of neurons \awn{in the second hidden layer}. Similar results we also found for even higher number\awn{s} of hidden layers and for other grid sizes. This indicates that \awn{a }BPNN with 1 hidden layer is \awn{preferable} for the application of BPNN to the online stage of FWIPS.
	\begin{figure}[!htb]
		\centering
		\includegraphics[width=0.75\columnwidth]{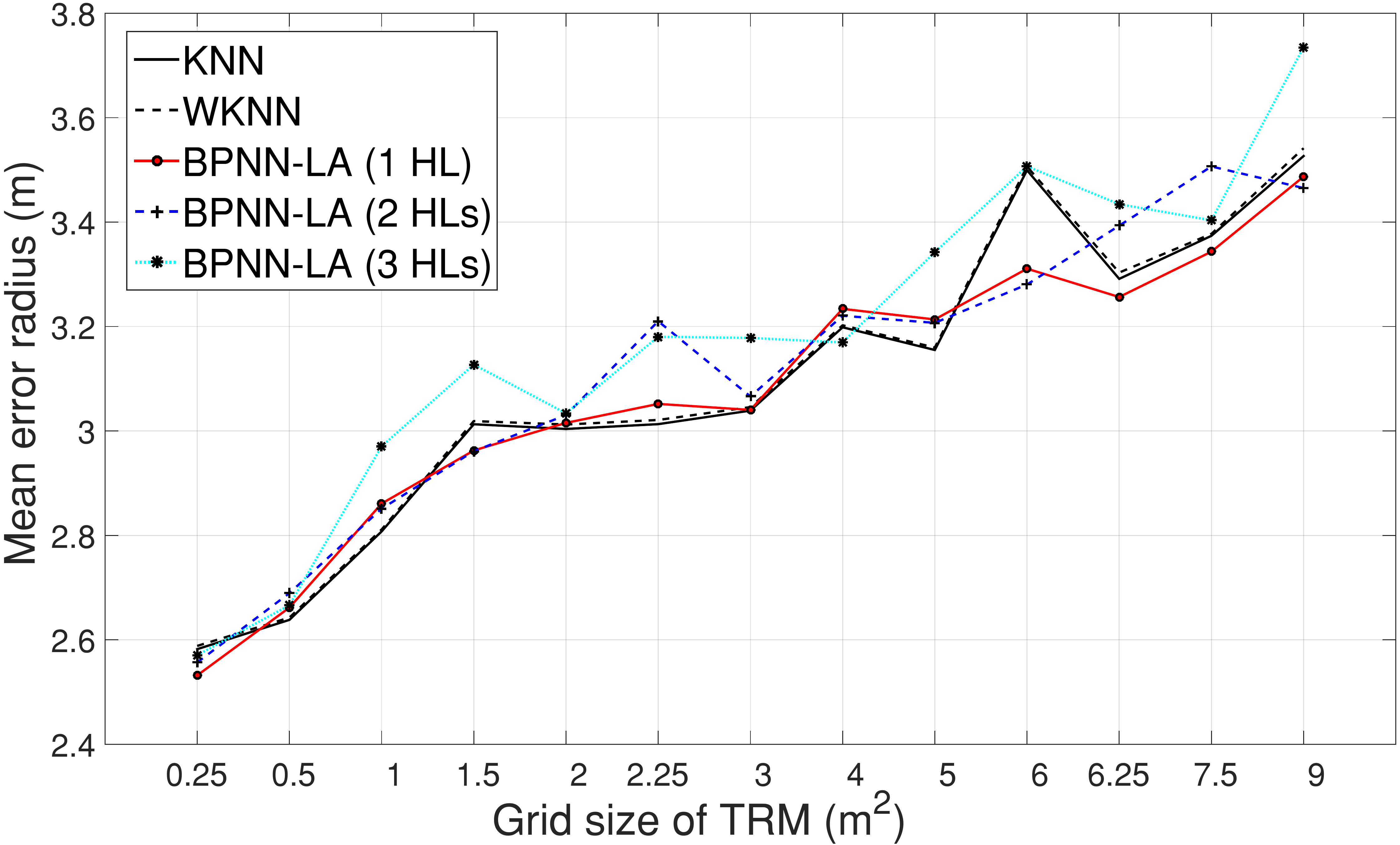}
		\caption{Comparison of MER with locally optimal number of nodes of BPNN-LA \awn{for different TRM grid sizes}}
		\label{fig_cmp_meanerror_optnode_bpnnla}
	\end{figure}
	\subsubsection{Locally optimal parameters of BPNN-LA for all TRMs} 
	\begin{table*}[!htb]
		\caption{Locally optimal parameters of BPNN-LA for several TRMs \awn{(GS: grid size; ER: error radius)}}
		\centering
		\begin{tabular}{p{0.03\columnwidth}|p{0.07\columnwidth}p{0.05\columnwidth}|p{0.05\columnwidth}p{0.05\columnwidth}|p{0.07\columnwidth}p{0.12\columnwidth}p{0.12\columnwidth}|p{0.1\columnwidth}p{0.07\columnwidth}p{0.05\columnwidth}|p{0.14\columnwidth}p{0.07\columnwidth}p{0.05\columnwidth}}
			\hline
			{{\shortstack{GS}}}&\multicolumn{2}{c|}{\shortstack{ER of $k$NN ($\mathrm{m}$)}}&\multicolumn{2}{c|}{\shortstack{ER of W$k$NN ($\mathrm{m}$)}}&\multicolumn{3}{c|}{\shortstack{ER of HL1}}&\multicolumn{3}{c|}{\shortstack{ER of HL2}}&\multicolumn{3}{c}{\shortstack{ER of HL3}}\\
			\centering
			•& {Mean}& {Std} &{Mean}& {Std} & $ \Gamma_1 $& {Mean}& {Std} & $ \Gamma_1 $, $ \Gamma_2 $& {Mean}& {Std} & $ \Gamma_1 $, $ \Gamma_2 $, $ \Gamma_3 $ & Mean & Std\\
			\hline
			0.25&2.58&1.49&2.59&1.49&21&2.53$\pm$0.04&1.49$\pm$0.13&18,15&2.56&1.59&23,12,16&2.57&1.58\\
			0.50&2.64&1.56&2.64&1.57&13&2.66$\pm$0.06&1.60$\pm$0.12&30,16&2.69&1.67&11,6,16&2.67&1.67\\
			1.00&2.81&1.68&2.81&1.69&6&2.86$\pm$0.09&1.71$\pm$0.30&3,16&2.85&1.80&7,24,27&2.97&1.97\\
			1.50&3.01&1.89&3.02&1.89&5&2.96$\pm$0.10&1.73$\pm$0.14&4,7&2.96&1.86&12,26,19&3.13&2.05\\
			2.00&3.00&1.85&3.01&1.85&6&3.02$\pm$0.10&1.89$\pm$0.31&4,13&3.03&2.00&4,9,29&3.03&1.98\\
			2.25&3.01&1.87&3.02&1.88&3&3.05$\pm$0.09&1.65$\pm$0.11&20,12&3.21&2.16&7,14,12&3.18&2.17\\
			3.00&3.04&1.88&3.05&1.90&4&3.04$\pm$0.11&1.76$\pm$0.19&5,14&3.07&2.06&6,25,23&3.18&2.09\\
			4.00&3.20&1.99&3.20&1.99&3&3.23$\pm$0.11&1.84$\pm$0.25&2,14&3.22&1.97&2,5,3&3.17&1.84\\
			5.00&3.16&2.04&3.16&2.04&2&3.21$\pm$0.14&1.74$\pm$0.11&2,8&3.21&2.03&3,23,26&3.34&2.17\\
			6.00&3.50&2.19&3.51&2.19&2&3.31$\pm$0.12&1.86$\pm$0.12&2,8&3.28&2.16&3,27,26&3.51&2.41\\
			6.25&3.29&2.03&3.30&2.04&3&3.26$\pm$0.15&1.84$\pm$0.19&2,23&3.39&2.19&7,6,23&3.43&2.31\\
			7.50&3.37&2.05&3.38&2.05&2&3.34$\pm$0.13&1.79$\pm$0.16&7,7&3.51&2.31&3,5,26&3.40&2.23\\
			9.00&3.53&2.13&3.54&2.14&2&3.49$\pm$0.16&1.90$\pm$0.16&2,4&3.47&2.03&30,15,19&3.73&2.37\\
			\hline
		\end{tabular}
		\label{tab_bpnnla_para}
	\end{table*}
	\begin{figure}[!htb]
		\centering
		\includegraphics[width=0.75\columnwidth]{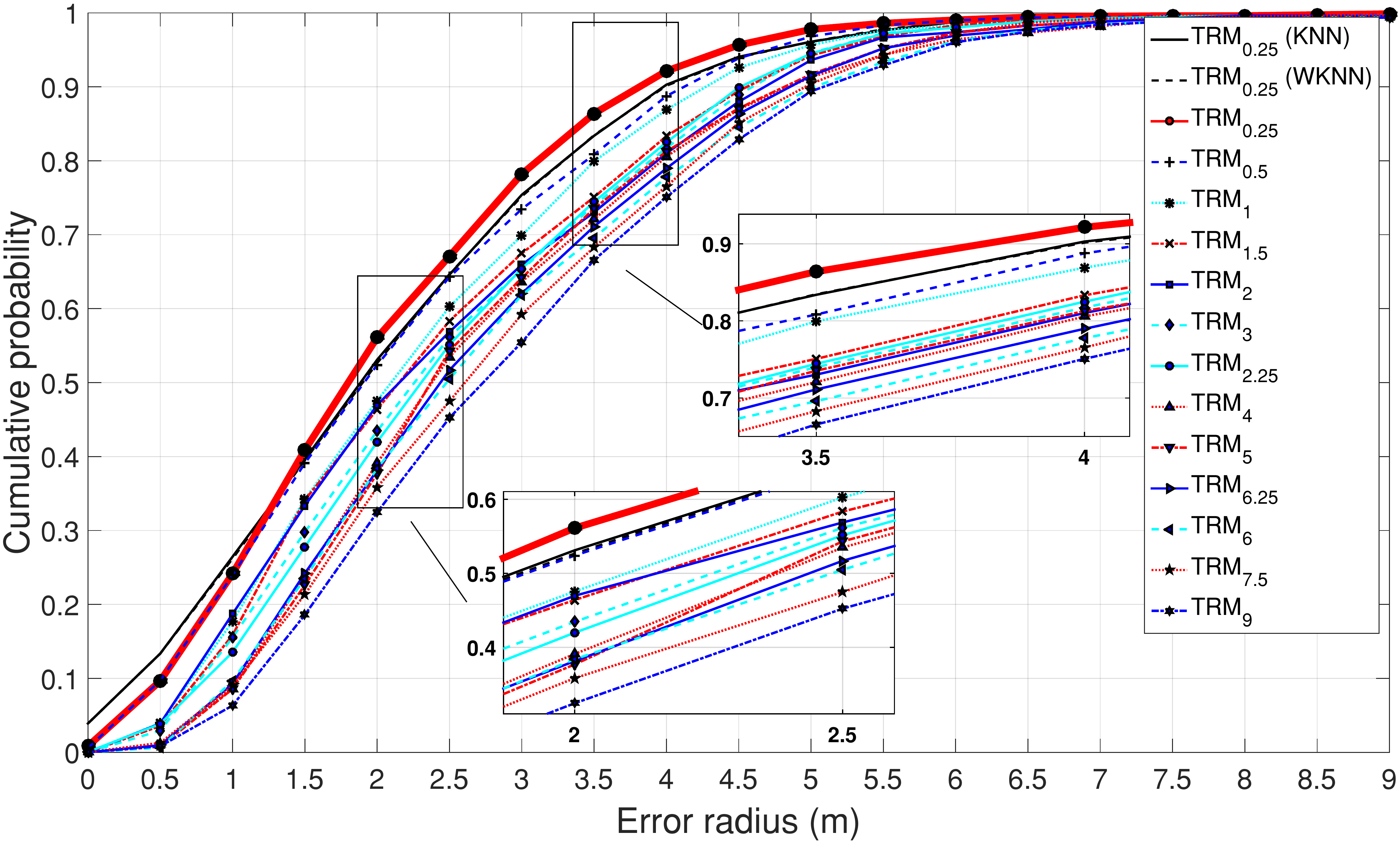}
		\caption{Cumulative probability of positioning \awn{error} for BPNN-LA}
		\label{fig_cumulative_positioning_accuracy_bpnnla}
	\end{figure}

	We report the locally optimal parameters of BPNN-LA with several different TRMs and 3 different numbers of hidden layers in Table \ref{tab_bpnnla_para}. For this analysis, we extracted 13 different TRMs from $\mathrm{ORM}_{0.25}$ with grid size \awn{varying} from 0.25\awn{\,}$\mathrm{m}^2$ to 9\awn{\,}$\mathrm{m}^2$. In the table, we find several patterns: (i) with a given BPNN-LA, for example HL1, the locally optimal number of neurons is proportional to the number of points in the TRM: the larger \awn{the }number of RPs in the TRM the bigger the locally optimal number of neurons. This pattern is also shared by HL2 and HL3, especially the locally optimal number of neurons of the last hidden layer. One explanation for this pattern is that smaller grid size of the TRM preserves more information of the nonlinearity of the \awn{underlying }RM, which is a key factor to determining the required number of neurons\cite{Larochelle2009}. (ii) With a specific TRM, HL1 achieves smaller MER as well as standard deviation than HL2 and HL3. This pattern is caused by \awn{increasing influence of }the random initialization of weights and biases \awn{with} increasing \awn{number }of hidden layers \cite{Larochelle2009}.
	
	\subsubsection{Influence of \awn{$\Lambda$}}
	Comparing BPNN-LA with 3 different hidden layers using the locally optimal number of neurons to $k$NN and W$k$NN in terms of the mean error radius, as shown in Fig.\awn{\,}\ref{fig_cmp_meanerror_optnode_bpnnla},  we can conclude: (i) HL1 outperforms HL2 and HL3 for almost all the analyzed grid sizes of \awn{the }TRM. This tendency is consistent with the results reported in \cite{Brunato2005}. (ii) \awn{$k$NN} achieves slightly smaller MER than \awn{W$k$NN in this example}. (iii) Comparing HL1 to W$k$NN, HL1 yields comparable performance with all grid sizes but in the case of particularly large grid size (larger than 6.25\awn{\,}$\mathrm{m}^2$) HL1 has slightly better performance than W$k$NN.
	\subsubsection{Cumulative positioning accuracy of BPNN-LA}
	In Fig.\awn{\,}\ref{fig_cumulative_positioning_accuracy_bpnnla}, we present the positioning accuracy of BPNN-LA with 1 hidden layer and the respective locally optimal number of neurons for all tested TRMs as well as that of $k$NN and W$k$NN as empirical \awn{distribution} function\awn{s}. With $\mathrm{TRM}_{0.25}$, BPNN-LA outperforms the other \awn{\,solutions and is even} better than $k$NN and W$k$NN with the same TRM. For BPNN-LA HL1 about 68\% of the errors are below 2.5\awn{\,}$\mathrm{m}$. Over 99\% of the estimated locations are within an error radius of 8 $\mathrm{m}$ which is accurate enough for room level positioning an\awn{d} ILBSs.
	\subsubsection{Computational complexity of BPNN-LA}
	Using the dimension of \awn{the }RSS \awn{vectors }(${N}$), the number of RPs (${M}$), hidden layers (${\Lambda}$) and neurons for each hidden layer ($\boldsymbol{\Gamma} = \{\Gamma_{\mathrm{in}}, \Gamma_1, \cdots, \Gamma_{\Lambda},\Gamma_{\Lambda+1}\}$), we can assess the computational complexity for the location estimation per request (i.e.\awn{,} one position required) during the online stage. \awn{For} $k$NN the computational complexity is $\mathcal{O}(k{N}{M})$. \awn{For} the generalization of BPNN-LA the computational complexity is $\mathcal{O}({\max\{\boldsymbol{\Gamma}, {N}\}}^2\awn{\Lambda})$ with the assumption that the \awn{evaluation of the} activation functions is negligible. These two computational complexities are comparable. The latter is smaller in the case of large number of RPs (\awn{i.e.,} ${M}\gg{N}$). Therefore, BPNN-LA also gains online computational efficiency in this case.

	\subsection{BPNN based radio map construction}\label{sec_4_3}
	
	Now we assume that the BPNN is used to construct a \awn{reconstructed} radio map RRM of grid size $G_R$ starting from a given radio map TRM of grid size $G_S$ with the purpose of using \awn{the }RRM for subsequent position estimation within an FWIPS. The underlying idea is that the TRM could result directly from sampling the RSS at a certain (moderate) number of RPs and could be converted into a denser radio map RRM (i.e.\awn{,} $G_R<G_S$) which ideally yields higher accuracy of the estimated positions than \awn{the }TRM. Higher accuracy could potentially even be obtained if $G_R \geq G_S$. When using BPNN-RM for this radio map reconstruction the accuracy of the positions finally obtained depends on the FLA, the quality of the measurements, on $G_R$, $G_S$, the number $\Lambda$ of hidden layers, and the numbers $\Gamma_1,\,\Gamma_2,\,\dots,\,\Gamma_\Lambda$ of nodes within the hidden layers. In this section we investigate this relationship for $k$NN and W$k$NN analyzing whether BPNN-RM can be used to increase the quality of the position estimation, in particular for the densification case which would be attractive because it would help to reduce the workload associated with radio map generation. \awn{Of course} it is possible to \awn{also use BPNN instead of $k$NN or W$k$NN for location estimation, as discussed in the previous section.} \awn{However, we will not} focus on this implementation herein.  
	
	We first analyze the situation for a small subset of free parameters and then generalize by calculating and discussing the locally optimal parameters for a variety of cases.
	
	\subsubsection{Locally optimal parameter\awn{s with} $\mathrm{TRM}_{7.5}$}
	\begin{figure}[!htb]
		\begin{center}
			\includegraphics[width=0.8\columnwidth,draft=false]{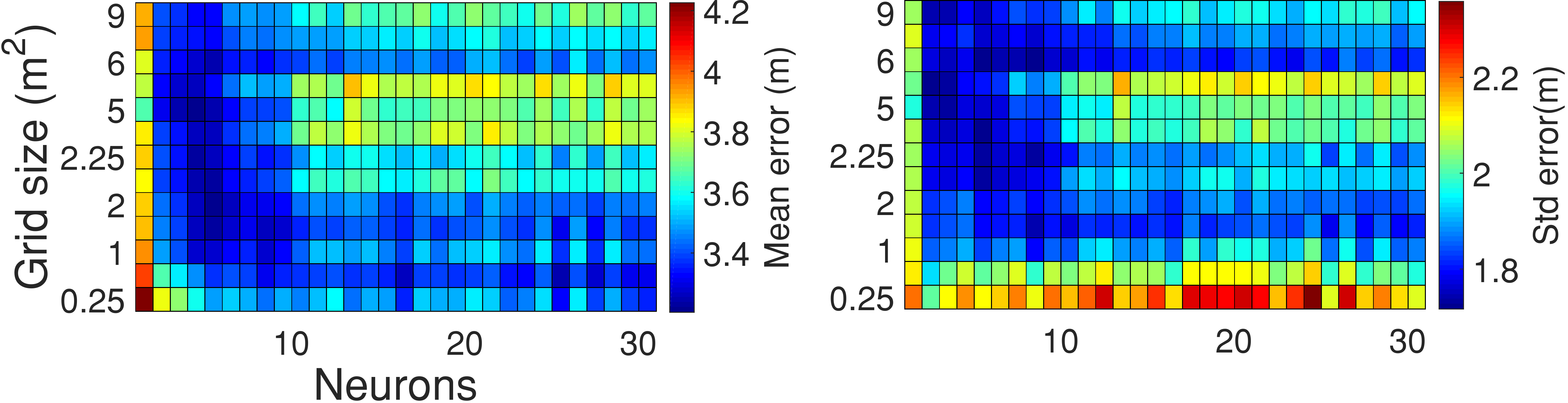}
			\label{fig_bpnnrm_knn_hl1_all}
		\end{center}
		\caption{Mean error of $k$NN using BPNN-RM with $\mathrm{TRM}_{7.5}$ (1 hidden layer)}
		\label{fig_error_of_bpnnrm_7_5}
	\end{figure}
	\begin{table*}[!t]
		\caption{Locally optimal parameters of BPNN-RM for selected combinations of TRM (i.e. $ G_S $) and RRM(i.e. $ G_S $)}
		\centering
		\begin{tabular}{p{0.03\columnwidth}p{0.03\columnwidth}|p{0.05\columnwidth}p{0.05\columnwidth}|p{0.03\columnwidth}p{0.12\columnwidth}p{0.12\columnwidth}|p{0.1\columnwidth}p{0.05\columnwidth}p{0.05\columnwidth}|p{0.14\columnwidth}p{0.05\columnwidth}p{0.05\columnwidth}}
			\hline
			$G_S$&$G_R$&\multicolumn{2}{c|}{\shortstack{ER of $k$NN}}&\multicolumn{3}{c|}{\shortstack{ER of $k$NN with RRM (1 layer)}}&\multicolumn{3}{c|}{\shortstack{ER of $k$NN with RRM (2 layers)}}&\multicolumn{3}{c}{\shortstack{ER of $k$NN with RRM (3 layers)}}\\
			•&•&{Mean}& {Std} & $ \Gamma_1 $& {Mean}& {Std} &$ \Gamma_1 $, $ \Gamma_2 $& {Mean}& {Std} &$ \Gamma_1 $, $ \Gamma_2 $, $ \Gamma_3 $&Mean&Std\\
			\hline
			0.25&0.25&2.58&1.49&18&3.19$\pm$0.15&2.16$\pm$0.19&29,28&2.92&1.81&28,27,23&2.88&1.76\\
			1.00&1.00&2.81&1.68&28&2.83$\pm$0.08&1.56$\pm$0.11&26,12&2.79&1.55&16,28,11&2.80&1.56\\
			2.00&2.00&3.00&1.85&26&2.90$\pm$0.08&1.62$\pm$0.08&9,7&2.93&1.72&6,25,23&2.90&1.67\\
			4.00&4.00&3.20&1.99&10&3.08$\pm$0.13&1.63$\pm$0.09&7,20&3.23&1.66&2,17,9&3.30&1.76\\
			6.00&6.00&3.50&2.19&15&3.29$\pm$0.16&1.72$\pm$0.07&3,23&3.35&1.75&2,29,6&3.43&1.84\\
			9.00&9.00&3.53&2.13&11&3.38$\pm$0.21&1.74$\pm$0.10&2,30&3.61&1.89&20,22,2&3.55&1.79\\
			0.25&4.00&2.58&1.49&19&3.03$\pm$0.06&1.76$\pm$0.11&30,16&2.95&1.61&12,26,16&2.94&1.59\\
			1.00&4.00&2.81&1.68&19&3.00$\pm$0.08&1.64$\pm$0.08&30,11&2.94&1.63&12,10,16&2.93&1.62\\
			2.00&4.00&3.00&1.85&25&3.04$\pm$0.11&1.61$\pm$0.07&14,8&3.00&1.64&17,7,10&3.02&1.66\\
			6.00&4.00&3.50&2.19&6&3.37$\pm$0.27&1.77$\pm$0.15&3,23&3.56&1.87&2,29,6&3.53&1.85\\
			9.00&4.00&3.53&2.13&2&3.62$\pm$0.50&1.92$\pm$0.35&2,6&3.42&1.80&5,11,13&3.61&1.91\\
			\hline
		\end{tabular}
		\label{tab_bpnnrm_para}
	\end{table*}
	The determination of the locally optimal parameters of BPNN-RM are determined using the same approach \awn{as for} BPNN-LA \awn{above}: for a give\awn{n} pair, $G_S$ and $G_R$, of grid sizes, a BPNN is trained using $\Lambda$ hidden layers\awn{, and} $\Gamma_1,\Gamma_2,\cdots\Gamma_{\Lambda}$ nodes; then \awn{the }VDS (testing point\awn{s} $\mathbf{l}^{(t)}$ and the corresponding RSS vectors $\mathbf{s}^{(t)}$) are used to calculate the positioning error at each $\mathbf{l}^{(t)}$. Repeating this for a variety of parameters the ones yielding the minimum MER are determined as locally optimal ones. We first present an example with $\mathrm{TRM}_{7.5}$ (i.e.\awn{,} $ G_S $=7.5\,$\mathrm{m}^2$), with 13 different grid sizes $G_R$, using BPNN-RM with 1 hidden layer whose number of neurons varies from 1 to 30. The results are visualized in Fig.\awn{\,}\ref{fig_error_of_bpnnrm_7_5}.
	
	The two \awn{plots} in Fig.\awn{\,}\ref{fig_error_of_bpnnrm_7_5} depict the MER (left) and \awn{the }standard deviation (right) of the error radius \awn{obtained using $k$NN\footnote{The results obtained using W$k$NN instead of $k$NN are virtually identical and not shown therefore.},} in terms of different number of neurons as well as different reconstruction grid sizes\awn{. They} show that the optimal number of neurons to achieve the minimum MER is \awn{mostly }consistent with the \awn{number }\awn{yielding} minimum standard deviation. According to the MER in the figure the locally optimal number of neurons for $\mathrm{RRM}_{7.5}$ is {2} in the case of \awn{densification from }$\mathrm{TRM}_{7.5}$ \awn{using a }BPNN-RM with 1 hidden layer.

	\subsubsection{Locally optimal parameter\awn{s} of BPNN-RM for \awn{a }variety of grid sizes}
	\awn{Using the extensive numerical simulations, as before, we have determined the locally optimal number of neurons for RRM generation as judged by the MER after positioning with $k$NN using the RRM. The results are given in Table \ref{tab_bpnnrm_para} for selected pairs of grid sizes $ G_S $ and $ G_R $ of the radio maps and for the selected numbers of hidden layers.} In the case that TRM and RRM have the same grid size (i.e.\awn{,} $G_S=G_R$), BPNN-RM with 1 hidden layer outperforms BPNN-RM with 2 or 3 hidden layers in terms of the MER for the grid sizes from 2\,$\mathrm{m}^2$ to 9\,$\mathrm{m}^2$. \awn{The location accuracy obtained using the RRM is also comparable to the one obtained using the TRM directly.} \awn{In a few cases, the results are slightly better with higher number of hidden layers. However, figuring in the uncertainty of the empirical results the benefit is not significant. So, we conclude that 1 hidden layer with an optimized number of nodes is sufficient.}
	
	\subsubsection{Comparison \awn{of} the performance \awn{related to }TRM and RRM}
	\begin{figure}[!htb]
		\begin{center}
			\includegraphics[width=0.8\columnwidth,draft=false]{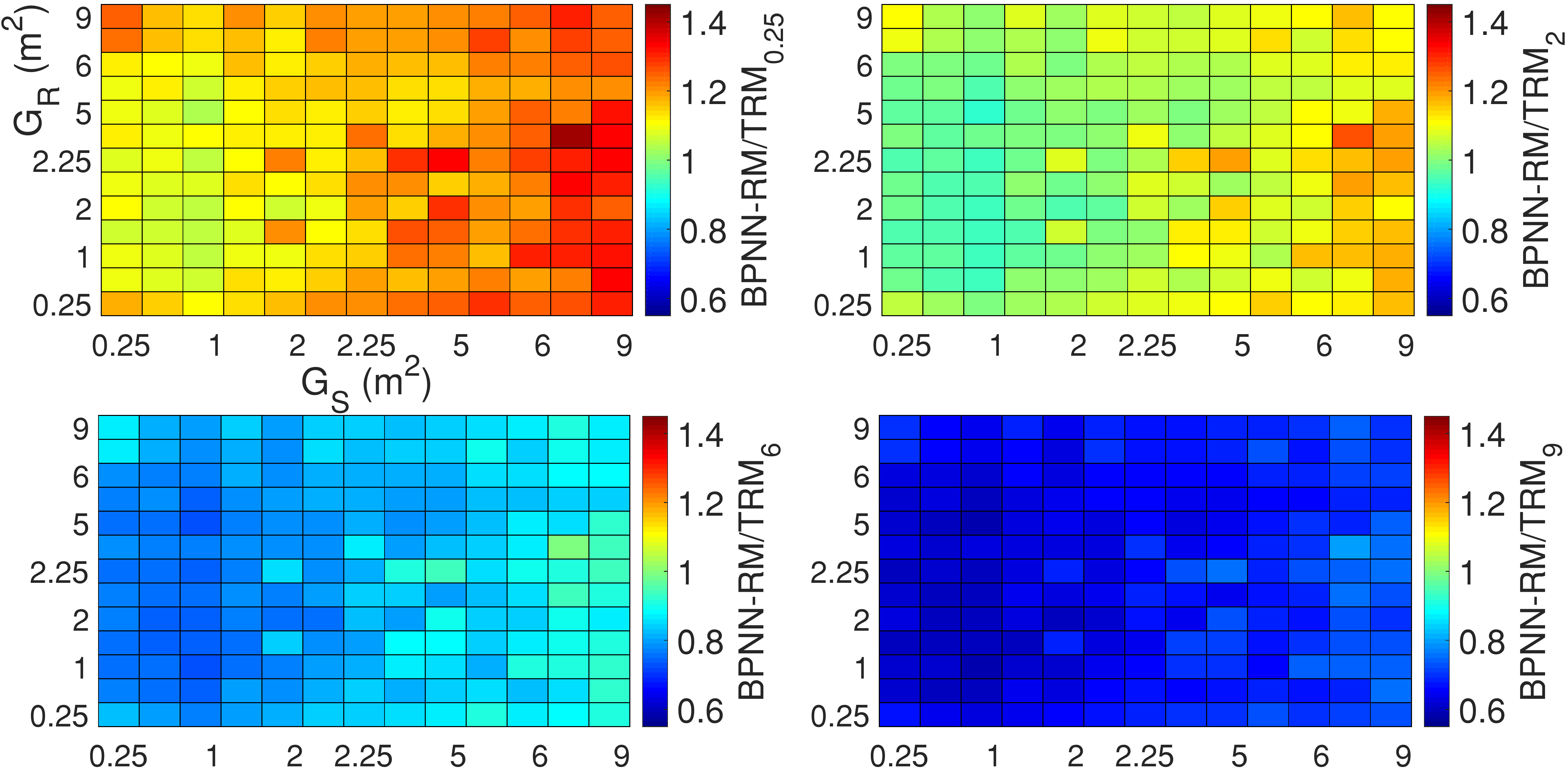}
			\label{fig_bpnnrm_trm_rrm_knn}
		\end{center}
		\caption{Comparison of between ORM and varies TRM w.r.t. MER using $k$NN}
		\label{fig_cmp_bpnnrm_trm_rrm}
	\end{figure}
	
	With the proposed BPNN-RM we expect to reduce the workload of the radio map construction while maintaining the positioning accuracy. Therefore, we present a comparison of the MER for several different grid sizes of TRM and RRM using $k$NN in Fig.\awn{\,}\ref{fig_cmp_bpnnrm_trm_rrm}. We can draw several conclusions from the figure: (i) $k$NN with RRM grid sizes $ G_R $ from 0.5 $\mathrm{m}^2$ to 5 $\mathrm{m}^2$, trained from a TRM with grid size $ G_S=\mathrm{1\ m}^2 $ achieves comparable performance to $ k $NN with an ORM of {0.25} $\mathrm{m}^2$ grid size. The maximal grid size $ G_S $ of \awn{the }TRM with which we obtained  comparable MER \awn{as }with an ORM of {0.25} $\mathrm{m}^2$ grid size is 2 $\mathrm{m}^2$. This means that only 1/8 of the workload for \awn{radio map} generation is required when reconstructing the radio map for $ k $NN using BPNN-RM instead of using the ORM directly for $ k $NN. (ii) Comparing the BPNN-RM results to $\mathrm{TRM}_{2}$, $\mathrm{TRM}_{6}$ and $\mathrm{TRM}_{9}$, the reduction of MER is up to 10\%, 20\% and 40\%\awn{,} respectively. BPNN-RM with W$k$NN leads to similar conclusions.
	
	\subsubsection{Cumulative \awn{error probability} of $k $NN and W$ k $NN using RRM}
	
	\begin{figure}[!htb]
		\begin{center}
			\subfigure[Cumulative \awn{error }probability \awn{for} $k$NN with RRM]{
				\includegraphics[width=0.75\columnwidth,draft=false]{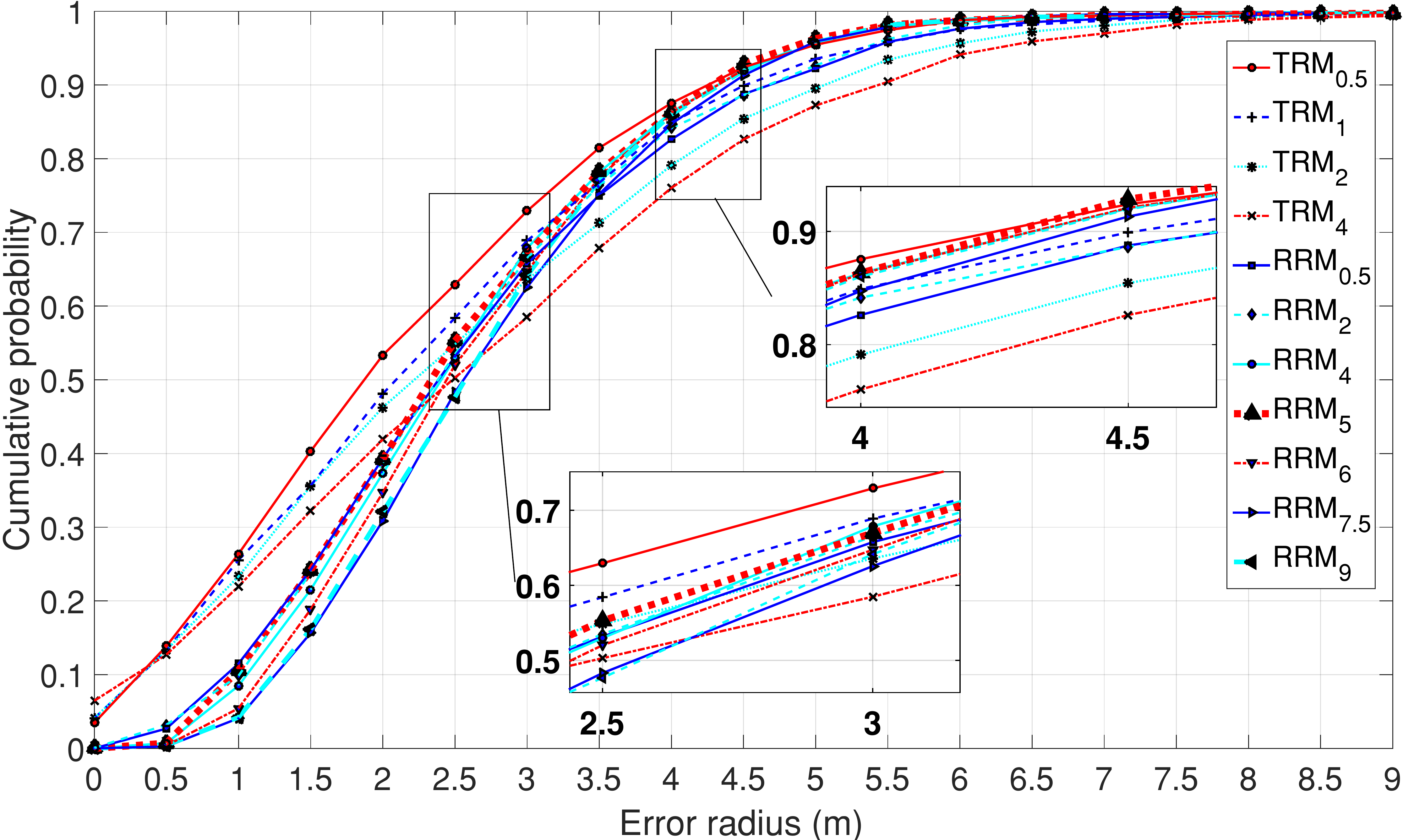}
				\label{fig_cumu_bpnnrm_knn}}
			\subfigure[Cumulative \awn{error }probability \awn{for} W$k$NN with RRM]{
				\includegraphics[width=0.75\columnwidth,draft=false]{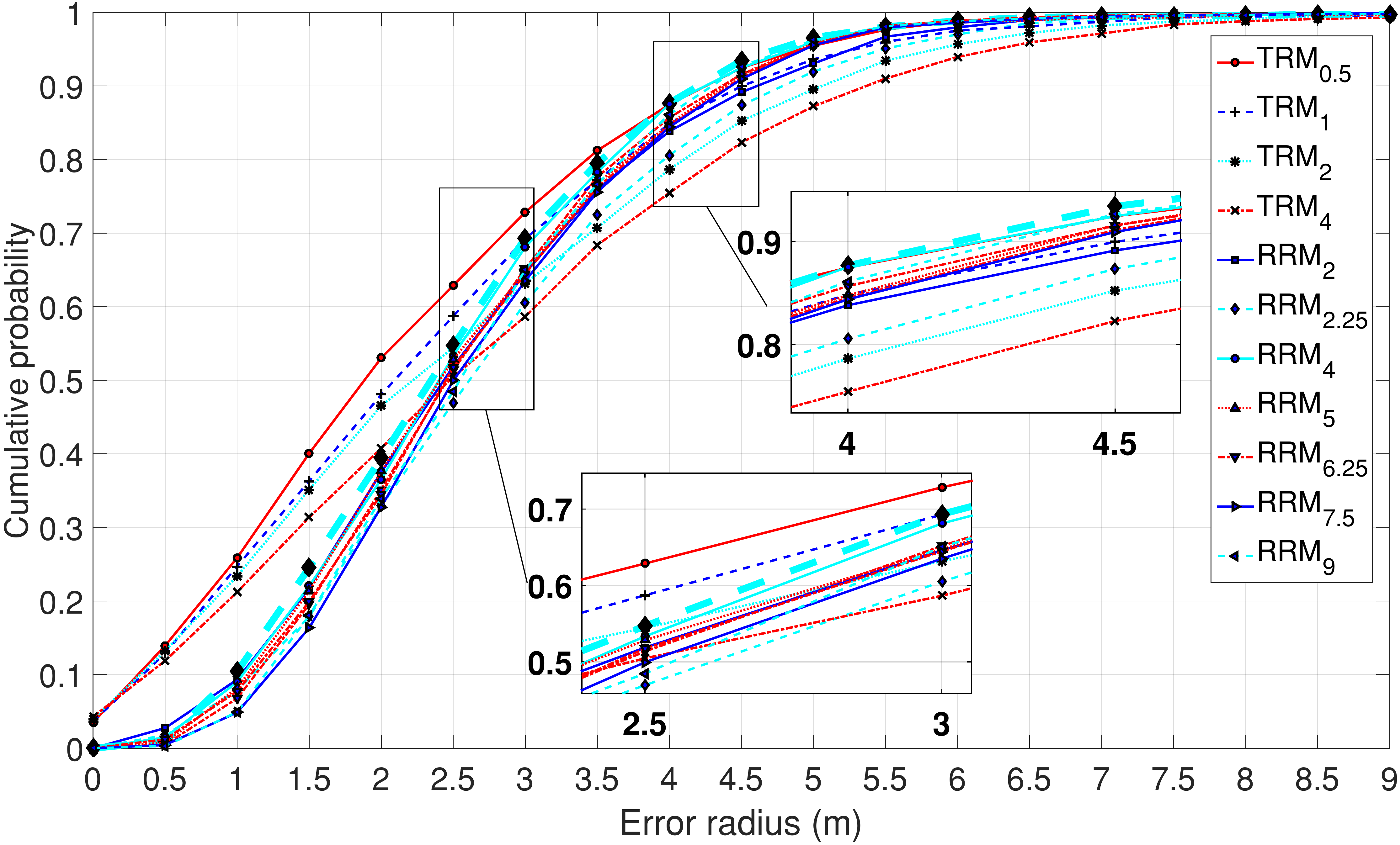}
				\label{fig_cumu_bpnnrm_wknn}}
		\end{center}
		\caption{Cumulative probability of $ k $NN and W$ k $NN with RRM }
		\label{fig_cumu_bpnnrm_knn_wknn}
	\end{figure}
	In this section, we compare the cumulative \awn{error probability of the positions estimated using $k$NN and W$k$NN for} 4 different TRMs and 6 RRMs which are reconstructed using the trained BPNN-RM from $\mathrm{TRM}_4$. As shown in Fig.\awn{\,}\ref{fig_cumu_bpnnrm_knn_wknn}, we \awn{find} that: (i) From 75\% to over 85\% of the errors are within 4 $ \mathrm{m} $. (ii) The positioning accuracy of both $k$NN and W$k$NN using the selected RRMs \awd{are}\awn{is} higher than using the respective TRM \awn{with equal grid size}\awd{, i.e., ${\mathrm{TRM}_4}$} when \awn{considering errors}\awd{the error radius is} larger than 2.5\,$\mathrm{m}$. (iii) The RRM\awn{s} \awd{which achieves}\awn{yielding} the best performance with $k$NN and W$k$NN \awd{is}\awn{are} $\mathrm{RRM}_5$ and $\mathrm{RRM}_{2.25}$\awn{;} 92\% and 95\% \awd{their error radius is}\awn{of the errors are} smaller than 4.5\,$\mathrm{m}$ \awn{when using them}. This positioning accuracy is better than \awn{the one obtained using directly the} $\mathrm{TRM}_{0.5}$\awd{ with $k$NN and W$k$NN} \awn{i.e., a much denser radio map associated with much higher workload for construction}. (iv) \awd{As for the reduction of workload required to build the TRM,}BPNN-RM reduces \awn{the workload for creating the radio map by almost}\awd{at least} \awd{87.5}\awn{90}\% \awd{of the workload by using }\awn{when collecting only the data required for} $\mathrm{TRM}_{4}$ \awn{instead of $\mathrm{TRM}_{0.5}$ and still obtaining better results by converting the $\mathrm{TRM}_{4}$ into a RRM with a grid size of e.g., 2.25\,$\mathrm{m}^2$}\awd{to obtain better performance (i.e. higher probability of the error radius within 4.5 $\mathrm{m}$) than $\mathrm{TRM}_{0.5}$}. This improvement \awd{gains}\awn{results} from the capability of \awn{the }BPNN to filter the noise in the measured signal strength\awn{s used for RM generation}.
	\section{Conclusion}
	The authors propose a scenario to apply BPNN to both stages of FWIPS: BPNN-LA \awn{for localization in the online stage }and BPNN-RM \awn{for radio map reconstruction in the offline stage}. BPNN-LA with 1 hidden layer (HL1) outperforms $k$NN, W$k$NN \awd{as well as}\awn{and} BPNN with multiple hidden layers in terms of the mean error radius\awd{ and cumulative positioning accuracy}. 90\% of \awn{the }positioning error\awn{s} \awn{are} within 4\,$\mathrm{m}$ using HL1 trained by the 0.25\,$\mathrm{m}^2 $ grid size radio map. As for BPNN-LA with multiple hidden layers (2 and 3 hidden layers analyzed herein), they \awd{achieved}\awn{yielded} higher mean error radius than \awd{that of} HL1. A trained BPNN-LA with \awd{HL1}\awn{one hidden layer} is \awn{computationally }more efficient during \awn{the }online stage than $k$NN and W$k$NN, especially in case of \awn{a }large number of reference points in the radio map.
	
	\awn{We have tested the benefit of} BPNN-RM \awn{for converting an originally sampled radio map into a reconstructed radio map of possibly different grid size.} \awn{In particular, the positioning errors after application of} both $k$NN and W$k$NN \awn{have been analyzed}. The reduction of the mean error radius \awn{attributed to RM reconstruction was found to be} up to 40\%. As for the reduction of \awn{the }workload required to build the RM, BPNN-RM reduces \awn{it by almost 90}\% \awn{since it allows} using $\mathrm{TRM}_{4}$ \awn{instead of $\mathrm{TRM}_{0.5}$ while still obtaining equal or even slightly} better performance.
	
	We expect that the results can be generalized to other fingerprinting based IPSs (e.g., IPSs based on Bluetooth, magnetic field) and WIPSs which are deployed in the heterogeneous RoI (e.g., the airports and big malls). We will investigate this further by exploring BPNN-LA/RM deep learning and assessing the performance for more general real world setting\awn{s} where RPs are not arranged in a regular grid and the RoI is not dominated by \awn{free} space such that the RSS-fields are more complex than in our examples. We expect BPNN to be even more beneficial in such cases while likely requiring more neurons in the hidden layers than in the case\awn{s} presented herein.  

	\section*{Acknowledgment}
	The authors thank Konrad Schindler for granting access to a high performance computing cluster at ETH for the purpose of the extensive numeric simulations used herein. The doctoral research of the first author is financed by the Chinese Scholarship Council (CSC).

	
	
	\bibliographystyle{IEEEtran}
	\bibliography{BPNN_ref}

\begin{thebibliography}{10}
\providecommand{\url}[1]{#1}
\csname url@samestyle\endcsname
\providecommand{\newblock}{\relax}
\providecommand{\bibinfo}[2]{#2}
\providecommand{\BIBentrySTDinterwordspacing}{\spaceskip=0pt\relax}
\providecommand{\BIBentryALTinterwordstretchfactor}{4}
\providecommand{\BIBentryALTinterwordspacing}{\spaceskip=\fontdimen2\font plus
\BIBentryALTinterwordstretchfactor\fontdimen3\font minus
  \fontdimen4\font\relax}
\providecommand{\BIBforeignlanguage}[2]{{%
\expandafter\ifx\csname l@#1\endcsname\relax
\typeout{** WARNING: IEEEtran.bst: No hyphenation pattern has been}%
\typeout{** loaded for the language `#1'. Using the pattern for}%
\typeout{** the default language instead.}%
\else
\language=\csname l@#1\endcsname
\fi
#2}}
\providecommand{\BIBdecl}{\relax}
\BIBdecl

\bibitem{Abowd}
\BIBentryALTinterwordspacing
G.~D. Abowd, ``What next, ubicomp?: Celebrating an intellectual disappearing
  act,'' in \emph{Proceedings of the 2012 ACM Conference on Ubiquitous
  Computing}, ser. UbiComp '12.\hskip 1em plus 0.5em minus 0.4em\relax New
  York, NY, USA: ACM, 2012, pp. 31--40. [Online]. Available:
  \url{http://doi.acm.org/10.1145/2370216.2370222}
\BIBentrySTDinterwordspacing

\bibitem{wifi_retail}
``{Wi-Fi Indoor Location in Retail Worth \$2.5 Billion by 2020},''
  {\url{https://www.abiresearch.com/press/wi-fi-indoor-location-retail-worth-25-billion-2020/}},
  2016, {[Online; Accessed: 2016-06-06]}.

\bibitem{Padmanabhan2000}
\BIBentryALTinterwordspacing
P.~B. Padmanabhan and V.~N., ``{RADAR: An in-building RF based user location
  and tracking system},'' \emph{Proceedings IEEE INFOCOM 2000. Conference on
  Computer Communications. Nineteenth Annual Joint Conference of the IEEE
  Computer and Communications Societies (Cat. No.00CH37064)}, vol.~2, no.~c,
  pp. 775--784, 2000. [Online]. Available:
  \url{http://research.microsoft.com/en-us/groups/sn-res/infocom2000.pdf}
\BIBentrySTDinterwordspacing

\bibitem{6934184}
E.~Lohan, K.~Koski, J.~Talvitie, and L.~Ukkonen, ``Wlan and rfid propagation
  channels for hybrid indoor positioning,'' in \emph{Localization and GNSS
  (ICL-GNSS), 2014 International Conference on}, June 2014, pp. 1--6.

\bibitem{liu2007survey}
H.~Liu, H.~Darabi, P.~Banerjee, and J.~Liu, ``Survey of wireless indoor
  positioning techniques and systems,'' \emph{Systems, Man, and Cybernetics,
  Part C: Applications and Reviews, IEEE Transactions on}, vol.~37, no.~6, pp.
  1067--1080, 2007.

\bibitem{6418880}
B.~Li, T.~Gallagher, A.~G. Dempster, and C.~Rizos, ``How feasible is the use of
  magnetic field alone for indoor positioning?'' in \emph{Indoor Positioning
  and Indoor Navigation (IPIN), 2012 International Conference on}, Nov 2012,
  pp. 1--9.

\bibitem{gigl2007analysis}
T.~Gigl, G.~J. Janssen, V.~Dizdarevi{\'c}, K.~Witrisal, and Z.~Irahhauten,
  ``Analysis of a uwb indoor positioning system based on received signal
  strength,'' in \emph{Positioning, Navigation and Communication, 2007.
  WPNC'07. 4th Workshop on}.\hskip 1em plus 0.5em minus 0.4em\relax IEEE, 2007,
  pp. 97--101.

\bibitem{hazas2006broadband}
M.~Hazas and A.~Hopper, ``Broadband ultrasonic location systems for improved
  indoor positioning,'' \emph{Mobile Computing, IEEE Transactions on}, vol.~5,
  no.~5, pp. 536--547, 2006.

\bibitem{mandal2005beep}
A.~Mandal, C.~V. Lopes, T.~Givargis, A.~Haghighat, R.~Jurdak, and P.~Baldi,
  ``Beep: 3d indoor positioning using audible sound,'' in \emph{Consumer
  communications and networking conference, 2005. CCNC. 2005 Second
  IEEE}.\hskip 1em plus 0.5em minus 0.4em\relax IEEE, 2005, pp. 348--353.

\bibitem{He2016}
S.~He and S.~H.~G. Chan, ``{Wi-Fi fingerprint-based indoor positioning: Recent
  advances and comparisons},'' \emph{IEEE Communications Surveys and
  Tutorials}, vol.~18, no.~1, pp. 466--490, 2016.

\bibitem{7362027}
K.~Majeed, S.~Sorour, T.~Al-Naffouri, and S.~Valaee, ``Indoor localization and
  radio map estimation using unsupervised manifold alignment with geometry
  perturbation,'' \emph{IEEE Transactions on Mobile Computing}, vol.~PP,
  no.~99, pp. 1--1, 2015.

\bibitem{Park2010}
\BIBentryALTinterwordspacing
J.-g. Park, B.~Charrow, D.~Curtis, J.~Battat, E.~Minkov, J.~Hicks, S.~Teller,
  and J.~Ledlie, ``{Growing an organic indoor location system},''
  \emph{Proceedings of the 8th international conference on Mobile systems
  applications and services MobiSys 10}, no. June, p. 271, 2010. [Online].
  Available: \url{http://portal.acm.org/citation.cfm?doid=1814433.1814461}
\BIBentrySTDinterwordspacing

\bibitem{Wu2013}
C.~Wu, Z.~Yang, Y.~Liu, and W.~Xi, ``{WILL: Wireless indoor localization
  without site survey},'' \emph{IEEE Transactions on Parallel and Distributed
  Systems}, vol.~24, no.~4, pp. 839--848, 2013.

\bibitem{Bernardos2010}
A.~M. Bernardos, J.~R. Casar, and P.~Tarr\'{i}o, ``{Real time calibration for
  RSS indoor positioning systems},'' \emph{2010 International Conference on
  Indoor Positioning and Indoor Navigation, IPIN 2010 - Conference
  Proceedings}, no. September, pp. 15--17, 2010.

\bibitem{Atia2013}
M.~M. Atia, A.~Noureldin, and M.~J. Korenberg, ``{Dynamic online-calibrated
  radio maps for indoor positioning in wireless local area networks},''
  \emph{IEEE Transactions on Mobile Computing}, vol.~12, no.~9, pp. 1774--1787,
  2013.

\bibitem{Pan2007}
\BIBentryALTinterwordspacing
S.~Pan, J.~Kwok, Q.~Yang, and J.~Pan, ``{Adaptive Localization in a Dynamic
  WiFi Environment through Multi-view Learning.}'' \emph{National conference on
  artificial Intelligence}, pp. 1108--1113, 2007. [Online]. Available:
  \url{http://www.aaai.org/Papers/AAAI/2007/AAAI07-176.pdf}
\BIBentrySTDinterwordspacing

\bibitem{6866898}
S.~Sorour, Y.~Lostanlen, S.~Valaee, and K.~Majeed, ``Joint indoor localization
  and radio map construction with limited deployment load,'' \emph{Mobile
  Computing, IEEE Transactions on}, vol.~14, no.~5, pp. 1031--1043, 2015.

\bibitem{Statistik2014}
\BIBentryALTinterwordspacing
B.~P. Statistik, ``{Neural network based indoor positioning technique in
  optical camera communication system},'' \emph{Katalog BPS}, vol. XXXIII,
  no.~2, pp. 81--87, 2014. [Online]. Available:
  \url{http://cid.oxfordjournals.org/lookup/doi/10.1093/cid/cir991$\backslash$}
\BIBentrySTDinterwordspacing

\bibitem{Wagner2012}
B.~Wagner, D.~Timmermann, G.~Ruscher, and T.~Kirste, ``{Device-free user
  localization utilizing artificial neural networks and passive RFID},''
  \emph{2012 Ubiquitous Positioning, Indoor Navigation, and Location Based
  Service, UPINLBS 2012}, 2012.

\bibitem{Xu2016}
\BIBentryALTinterwordspacing
J.~Xu, H.~Dai, and W.-h. Ying, ``{Multi-layer neural network for received
  signal strength-based indoor localisation},'' \emph{IET Communications},
  vol.~10, no.~6, pp. 717--723, 2016. [Online]. Available:
  \url{http://digital-library.theiet.org/content/journals/10.1049/iet-com.2015.0469}
\BIBentrySTDinterwordspacing

\bibitem{Soltani2013}
\BIBentryALTinterwordspacing
M.~M. Soltani, A.~Motamedi, and A.~Hammad, ``{Enhancing Cluster-based RFID Tag
  Localization using artificial neural networks and virtual reference tags},''
  \emph{International Conference on Indoor Positioning and Indoor Navigation},
  no. October, pp. 1--10, 2013. [Online]. Available:
  \url{http://ieeexplore.ieee.org/lpdocs/epic03/wrapper.htm?arnumber=6817886}
\BIBentrySTDinterwordspacing

\bibitem{Edel2015}
M.~Edel and E.~Koppe, ``{An advanced method for pedestrian dead reckoning using
  BLSTM-RNNs},'' \emph{2015 International Conference on Indoor Positioning and
  Indoor Navigation, IPIN 2015}, no. October, pp. 13--16, 2015.

\bibitem{Kulkarni2011}
\BIBentryALTinterwordspacing
S.~R. Kulkarni and G.~Harman, ``{Statistical learning theory: A tutorial},''
  \emph{Wiley Interdisciplinary Reviews: Computational Statistics}, vol.~3,
  no.~6, pp. 543--556, 2011. [Online]. Available:
  \url{http://onlinelibrary.wiley.com/doi/10.1002/wics.179/epdf}
\BIBentrySTDinterwordspacing

\bibitem{Larochelle2009}
H.~Larochelle, Y.~Bengio, J.~Louradour, and P.~Lamblin, ``{Exploring Strategies
  for Training Deep Neural Networks},'' \emph{Journal of Machine Learning
  Research}, vol.~1, pp. 1--40, 2009.

\bibitem{Brunato2005}
M.~Brunato and R.~Battiti, ``{Statistical learning theory for location
  fingerprinting in wireless LANs},'' \emph{Computer Networks}, vol.~47, no.~6,
  pp. 825--845, 2005.

\bibitem{Hagan1995}
\BIBentryALTinterwordspacing
M.~T. Hagan, H.~B. Demuth, and M.~H. Beale, ``{Neural Network Design},'' pp.
  1--1012, 1995. [Online]. Available:
  \url{http://books.google.ru/books?id=bUNJAAAACAAJ}
\BIBentrySTDinterwordspacing

\bibitem{69}
C.~Zhou, L.~Ma, and X.~Tan, ``Joint semi-supervised rss dimensionality
  reduction and fingerprint based algorithm for indoor localization,'' in
  \emph{Institute of Navigation (ION GNSS+2014), 27th International Technical
  Meeting of The Satellite Division Conference on}, September 2014, pp.
  3201--3211.

\end{thebibliography}
	%
	%
	%

\end{document}